\documentclass{article}

\usepackage{hyperref}
\usepackage{xspace}
\usepackage{booktabs}
\usepackage{amssymb}

\usepackage[accepted]{icml2021}

\icmltitlerunning{Learning Task Informed Abstractions}
\definecolor{mydarkblue}{rgb}{0,0.08,0.45}
\definecolor{es-blue}{rgb}{0,0.4,0.8}

\hypersetup{%
colorlinks=true,
linkcolor=mydarkblue,
citecolor=mydarkblue,
filecolor=mydarkblue,
urlcolor=mydarkblue}

\usepackage{xcolor}
\usepackage{xparse}
\usepackage{xpatch}
\usepackage{amsmath}
\usepackage{amsfonts}
\usepackage{mathtools}
\usepackage[capitalise,noabbrev,nameinlink]{cleveref}
\usepackage{tikz}
\usepackage{tikzscale}
\usepackage{graphicx}
\usepackage{enumitem}
\usepackage{tikzscale}
\usepackage{afterpage}
\usepackage{multicol}
\usepackage{etoolbox}
\usepackage{subcaption}
\usepackage{microtype}
\usepackage{booktabs}
\usepackage{multirow}
\usepackage{relsize}

\usetikzlibrary{positioning,arrows.meta,fit,calc,backgrounds,decorations.pathmorphing}

\makeatletter
\newcommand{\removeParBefore}{\ifvmode\vspace*{-\baselineskip}\setlength{\parskip}{0ex}\fi}
\newcommand{\removeParAfter}{\@ifnextchar\par\@gobble\relax}

\makeatother

\newcommand\defined{\doteq}

\DeclareMathOperator*{\argmax}{argmax}

\DeclareMathSymbol{\shortminus}{\mathbin}{AMSa}{"39}

\newlength{\philength}
\DeclarePairedDelimiterX{\RoundBrackets}[1]{(}{)}{#1}

\NewDocumentCommand{\pr}{ O{p} r() }{
  \def\prArg{#2}\patchcmd{\prArg}{|}{\mid}{}{}#1\RoundBrackets{\prArg}}
\NewDocumentCommand{\p}{ r() }{\pr[p](#1)}
\NewDocumentCommand{\q}{ r() }{\pr[q](#1)}
\NewDocumentCommand{\Normal}{ r() }{\pr[\operatorname{Normal}](#1)}
\NewDocumentCommand{\Cat}{ r() }{\pr[\operatorname{Cat}](#1)}
\NewDocumentCommand{\Bin}{ r() }{\pr[\operatorname{Bin}](#1)}
\NewDocumentCommand{\Beta}{ r() }{\pr[\operatorname{Beta}](#1)}
\NewDocumentCommand{\Bernoulli}{ r() }{\pr[\operatorname{Bernoulli}](#1)}
\NewDocumentCommand{\Dir}{ r() }{\pr[\operatorname{Dir}](#1)}

\newcommand{\E}[3][\big]{\mathbb{\operatorname{E}}_{#2}#1(#3#1)}

\newcommand{\KL}[3][\big]{\operatorname{KL}#1(#2\;#1\|\;#3#1)}
\newlength\widthE

\DeclareDocumentCommand{\pp}{ D<>{} r() }{
\ensuremath{\p<p_\phi><#1>(#2)}}
\DeclareDocumentCommand{\qp}{ D<>{} r() }{
\ensuremath{\p<q_\phi><#1>(#2)}}

\providecommand{\ourmethod}{\underline{\textbf{T}}ask \underline{\textbf{I}}nformed \underline{\textbf{A}}bstractions\xspace}

\begin{document}

\twocolumn[
\icmltitle{
Learning Task Informed Abstractions
}

\icmlsetsymbol{equal}{*}
\begin{icmlauthorlist}
\icmlauthor{Xiang Fu}{equal,mit}
\icmlauthor{Ge Yang}{equal,mit,iaifi}
\icmlauthor{Pulkit Agrawal}{mit,iaifi}
\icmlauthor{Tommi Jaakkola}{mit}
\end{icmlauthorlist}
\icmlaffiliation{mit}{MIT CSAIL}
\icmlaffiliation{iaifi}{IAIFI}
\icmlcorrespondingauthor{Xiang Fu}{xiangfu@csail.mit.edu}
\icmlkeywords{Machine Learning, ICML}
\vskip 0.3in
]

\printAffiliationsAndNotice{\icmlEqualContribution} 

\begin{abstract}
Current model-based reinforcement learning methods struggle when operating from complex visual scenes due to their inability to prioritize task-relevant features. To mitigate this problem, we propose learning \textbf{Task Informed Abstractions} (TIA) that explicitly separates reward-correlated visual features from distractors. For learning TIA, we introduce the formalism of Task Informed MDP (TiMDP) that is realized by training two models that learn visual features via cooperative reconstruction, but one model is adversarially dissociated from the reward signal. Empirical evaluation shows that TIA leads to significant performance gains over state-of-the-art methods on many visual control tasks where natural and unconstrained visual distractions pose a formidable challenge. Project page: \hyperlink{https://xiangfu.co/tia}{https://xiangfu.co/tia}
\end{abstract}
\section{Introduction}

\begin{figure}[t!]
    \centering
    \includegraphics[width=0.48\textwidth]{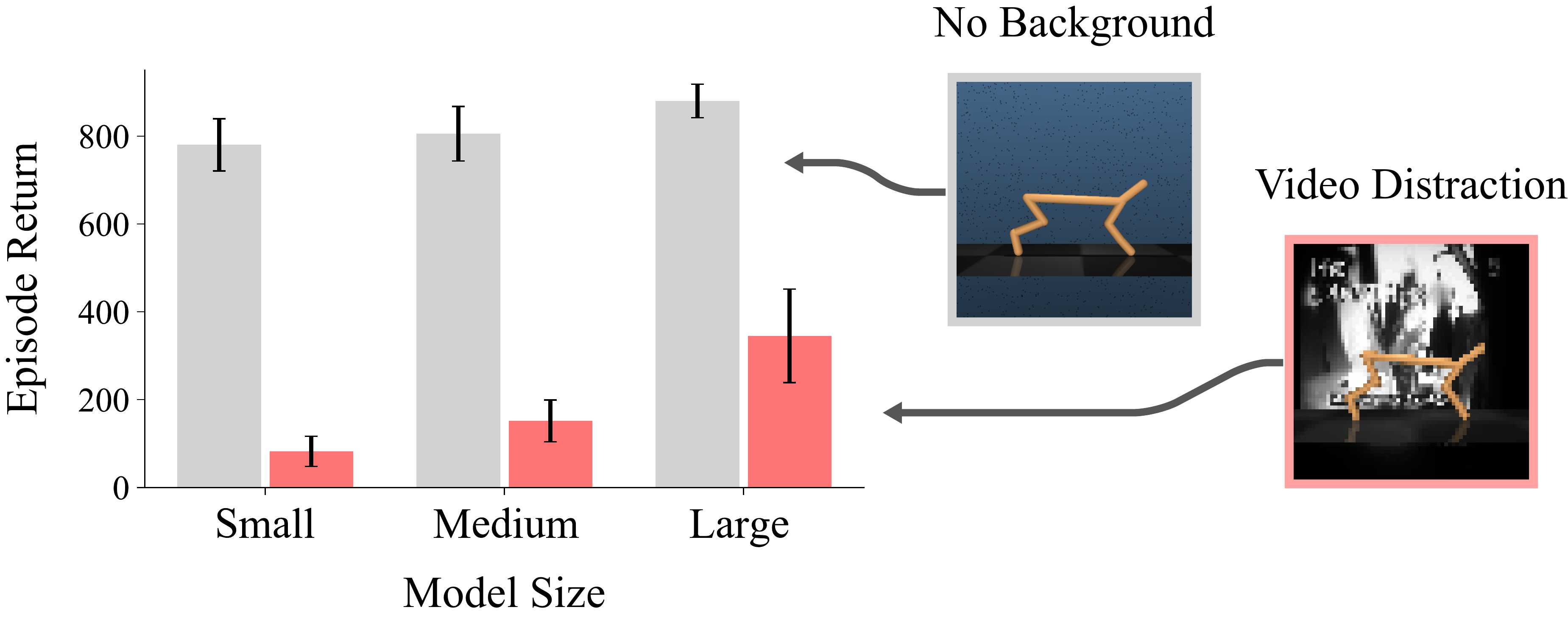}
    \caption{Comparison of the performance of a state-of-the-art model-based RL algorithm, \textit{Dreamer}, on two versions of the \textit{Cheetah Run} with vs. without visual distraction. Performance is reported for three models of increasing sizes (\(0.5\!\times\), \(1\!\times\), \(2\!\times\) of the original Dreamer). %
    Results show that even the smallest model has sufficient capacity to capture task-relevant features when observations are distractor-free~({\color{gray}gray}), but when the scene is complex ({\color{red}red}), task-irrelevant features consume most of the model capacity. Error bars indicate one standard deviation.
    }
    \label{fig:model_capacity}
\end{figure}

Consider results of a simple experiment reported in~\cref{fig:model_capacity}. We train a state-of-the-art model-based reinforcement learning algorithm~\cite{hafner2019dream} to solve two versions of the \textit{Cheetah Run} task~\citep{tassa2018dmc}: one with a simple and the other with a visually complex background~\cite{zhang2020learning}. For each version we train three model variants that contain \(0.5\!\times\) (small), \(1\!\times\) (medium) and \(2\!\times\) (large) parameters in their respective world models. The performance with the simple background is only marginally affected by model capacity indicating that even the smallest model has sufficient capacity to learn task-relevant features. With complex background, performance is much worse but increases monotonically with the model size. Because the amount of task-relevant information is unchanged between simple and complex background variants, these results demonstrate that excess model capacity is devoted to representing background information when learning from complex visual inputs. The background conveys no information about the task. It, therefore, interferes with the learning of task-relevant information by consuming model capacity. Here ``relevant" refers to features needed to predict the optimal actions, whereas ``irrelevant" refers to everything else that makes up the observation.

There are two main components of a model-based learner: (i) a forward dynamics model that predicts future events resulting from executing a sequence of actions from the current state and (ii) a reward predictor for evaluating possible future states. The policy performance depends critically on the prediction accuracy of the forward model, which is intimately tied to the feature space in which the future is predicted. Similar to the complex background version of \textit{Cheetah Run}, visual observations obtained in the real world are full of irrelevant information. Therefore, if there is no bias for learning ``task-relevant features", the model will try to predict all the information. In such a scenario spurious features would unnecessarily increase the sample complexity and necessitate the use of much larger models than necessary. Sometimes it can also result in training failure. 

Ways to learn good representations has been a major focus in model-based deep reinforcement learning. A popular choice is to reconstruct the input observations~\cite{kingma2014stochastic,kingma2014semi, watter2015embed,hafner2019dream}. Often these features are encouraged to be \textit{disentagled}~\cite{bengio2013representation,higgins2016beta,higgins2017darla} to identify distinct factors of variation. Since disentanglement simply re-formats the space, this disentangled feature space would still contain irrelevant information and will not address the central problem of learning task-relevant features. As our analysis shows, the incorporation of reward prediction loss is insufficient for producing a feature space that only contains task-relevant information. Reward supervision alone is shown insufficient for feature learning~\cite{yarats2019sample}. For instance, just knowing the center of mass of a humanoid moving forward is sufficient to predict the reward, whereas the knowledge of the full-body pose might be necessary to predict the optimal action. In a nutshell \textit{reconstruction captures too much information}, whereas \textit{reward-prediction captures too little}. Several works attempt to combine these two training signals~\cite{hafner2019dream,oh2017value} but struggle to learn in complex visual scenarios. Since the goal of the agent is to maximize the expected return, predicting the value function instead of one-step reward may aid in learning all the relevant information~\cite{silver2016predictron,oh2017value, schrittwieser2019mastering}. However, because the value function is often learned via bootstrapping, it may provide an unstable training signal.%

These challenges in learning task-relevant representations inspired several works to investigate feature learning methods that neither rely on reconstruction nor solely depend on rewards. One line of work biases the learned features to only capture controllable parts of the environment using an inverse model that predicts actions from a pair of states~\cite{agrawal2015learning,jayaraman2015learning,agrawal2016learning,pathak2017curiosity}, or using metrics such as  empowerment~\cite{klyubin2005empowerment,gregor2016variational}. To understand their shortcoming, consider the scenario of the arm pushing an object. Here both the arm and the object are controllable. While it is easy to capture the part that is directly controllable (e.g., the arm), capturing all controllable features (i.e., arm and the object) without imposing a reconstruction loss is non-trivial. Another idea that has shown promise is the bisimulation metric~\cite{ferns2011bisimulation, zhang2020learning}. Because supervision in bisimulation comes solely from rewards, it is subject to the same issues as mentioned earlier. Another possibility is to use contrastive learning~\cite{chen2020simple,oord2018representation}, but without additional constraints, these methods may not distinguish between relevant and irrelevant features. 

The ongoing discussion illustrates the fundamental challenge in learning task-relevant features: some objectives (e.g., reconstruction) capture too much information, whereas others (e.g., rewards, inverse models, empowerment) capture too little. Empirically we find that a weighted loss function that combines these objectives does not lead to task-relevant features (see \cref{fig:model_capacity}). In this work, we revisit feature learning by combining image reconstruction and reward prediction but propose to explicitly ``explain away" irrelevant features by constructing a cooperative two-player game between two models. These models, dubbed as task and distractor models, learn task-relevant ($s_t^+$) and irrelevant features ($s_t^-$) of the observation ($o_t$) respectively. Similar to prior work, we force the task model to learn task-relevant features ($s_t^+$) by predicting the reward. But unlike past work, we also force the distractor model to learn task-irrelevant features ($s_t^-$) via adversarial dissociation with the reward signal. However, both models cooperate to reconstruct $o_t$ by maximizing $p(o_t | s_t^+, s_t^-)$. 

Our method implements a Markov decision process (MDP) of a specific factored structure, which we call \textbf{Task Informed MDP (TiMDP)} (see \cref{fig:ti_mdp}). It is worth noting that TiMDP is structurally similar to the \textit{relaxed block MDP}~\citep{zhang2020invariant} formulation in partitioning the state-space into two separate components. However, \citep{zhang2020invariant} neither proposes a practical method for segregating relevant information nor provides any experimental validation of their framework in the context of learning from complex visual inputs. 
We evaluate our method on a custom ManyWorld environment, a suite of control tasks that specifically test the robustness of learning to visual distractions~\cite{zhang2020learning} and Atari games. The results convincingly demonstrate that our method, which we call \ourmethod (TIA), successfully learns relevant features and outperforms existing state-of-the-art methods. 
\section{Preliminaries}
\label{sec:preliminaries}

\paragraph{A Markov Decision Process} is represented as the tuple \(\langle\mathcal S, \mathcal{O}, \mathcal{A}, T, r, \gamma, \rho_0\rangle\) where \(\mathcal{O}\) is a high-dimensional observation space. \(\mathcal A\)  is the space of actions. \(\mathcal S\) is the state space. $\rho_0$ is the initial state distribution. \(r: \mathcal S \mapsto \mathbb R\) is the scalar reward. The goal of RL is to learn a policy \(\pi^*(a\vert\, s)\) that maximizes cumulative reward \(\mathcal J_\pi = \argmax_\pi \mathbb E\sum_t{\gamma^{t-1} r_t }\) discounted by \(\gamma\).

Our primary contribution is in the method for learning forward dynamics and is agnostic to the specific choice of the model-based algorithm. We choose to build upon the state-of-the-art method Dreamer~\cite{hafner2019dream}. The main components of this model are:
\begin{equation}
\begin{aligned}
\makebox[10em][l]{Representation model:} && &\pr[p_\theta](s_t  \,\vert\, o_t, s_{t-1},\, a_{t-1}) \\
\makebox[10em][l]{Observation model:} && &\pr[q_\theta](o_t \,\vert\, s_t) \\
\makebox[10em][l]{Transition model:} && &\pr[q_\theta](s_t\,\vert\,s_{t-1},\,a_{t-1})\\
\makebox[10em][l]{Reward model:} && &\pr[q_\theta](r_t\,\vert\,s_t)
\label{eq:dreamer_dynamics_model}
\end{aligned}
\end{equation}%

\paragraph{Model Learning} 
Dreamer~\cite{hafner2019dream} forecasts in a feature representation of images learned via supervision from three signals: (a) image reconstruction $\big[\mathcal{J}_{\mathrm{O}}^t \defined\ \ln\q(o_t|s_t) \big]$, (b) reward prediction $\big[\mathcal{J}_{\mathrm{R}}^t \defined\ \ln\q(r_t|s_t) \big]$ and (c) dynamics regularization $\Big[ \mathcal{J}_{\mathrm{D}}^t
\defined -\beta \mathrm{KL} \big(\p(s_t|s_{t-1},a_{t-1},o_t) \big\Vert \q(s_t|s_{t-1},a_{t-1})\big)\Big]$. The overall objective is: 
\begin{equation}
\mathcal{J}_{\mathrm{Dreamer}}
  \defined\ \mathbb{E}_{\tau} \bigg[\sum_t \mathcal{J}_{\mathrm{O}}^t + \mathcal{J}_{\mathrm{R}}^t + \mathcal{J}_{\mathrm{D}}^t\bigg]
\label{eq:dreamer_loss}
\end{equation}%
optimized over the agent's experience $\tau$. To achieve competitive performance on Atari, a few modifications are required that are incorporated in the variant \textit{DreamerV2} described in~\cite{hafner2020mastering}.

\paragraph{Policy Learning} 
Dreamer uses the learned forward dynamics model to train a policy using an actor-critic formulation described below:
\begin{equation}
\begin{aligned}
\makebox[5em][l]{Action model:} && &a_\tau \sim\ \pr[q_\phi](a_\tau|s_\tau) \\
\makebox[5em][l]{Value model:} && &v_\psi(s_\tau) \approx \E{q(\cdot|s_\tau)}{\textstyle\sum_{\tau=t}^{t+H}\gamma^{\tau-t}r_\tau}
\label{eq:dreamer_action_value_model}
\end{aligned}
\end{equation}%
The action model is trained to maximize cumulative rewards over a fixed horizon \(H\). Both the action and value models are learned using imagined rollouts from the learned dynamics.  We refer the reader to \cite{hafner2019dream} for more details.
\section{Learning Task Informed Abstractions}
\label{sec:learning_timdp}

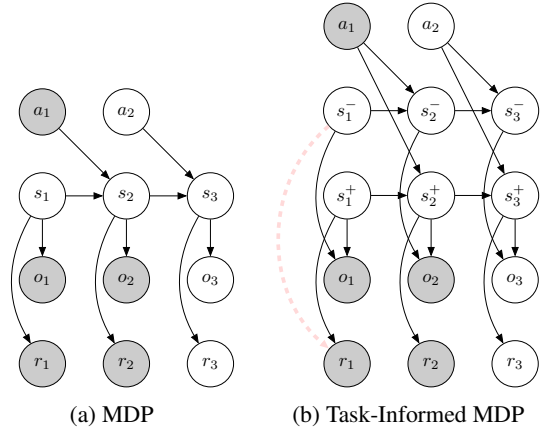
\begin{figure}[t!]
\begin{subfigure}[t]{0.23\textwidth}
\scalebox{0.7}{%
\usetikzlibrary{shapes.misc}
\begin{tikzpicture}[
  node distance=2em, auto,
  lat/.style={draw=black, circle, minimum size=2.5em},
  det/.style={draw=black, rectangle, minimum size=2.5em},
  obs/.style={circle, draw=black, fill=black!20, minimum size=2.5em},
  gen/.style={->, -{Stealth[length=.5em, inset=0pt]}},
  inf/.style={dashed, ->, -{Stealth[length=.5em, inset=0pt]}},
]
\node[obs, inner sep=.02em] (r1) {$r_1$};
\node[obs, right=of r1, inner sep=.02em] (r2) {$r_2$};
\node[lat, right=of r2, inner sep=.02em] (r3) {$r_3$};
\node[obs, above=of r1] (o1) {$o_1$};
\node[obs, above=of r2] (o2) {$o_2$};
\node[lat, above=of r3] (o3) {$o_3$};
\node[lat, above=of o1] (x1) {$s_1$};
\node[lat, above=of o2] (x2) {$s_2$};
\node[lat, above=of o3] (x3) {$s_3$};
\node[obs, above=of x1] (a1) {$a_1$};
\node[lat, above=of x2] (a2) {$a_2$};
\path (x1) edge[gen] node {} (o1);
\path (x1) edge[gen, bend right] node {} (r1);
\path (x2) edge[gen] node {} (o2);
\path (x2) edge[gen, bend right] node {} (r2);
\path (x3) edge[gen] node {} (o3);
\path (x3) edge[gen, bend right] node {} (r3);
\path (x1) edge[gen] node {} (x2);
\path (x2) edge[gen] node {} (x3);
\path (a1) edge[gen] node {} (x2);
\path (a2) edge[gen] node {} (x3);
\end{tikzpicture}}
\caption{MDP}
\label{fig:mdp}
\end{subfigure}%
\begin{subfigure}[t]{0.23\textwidth}
\scalebox{0.7}{%
\begin{tikzpicture}[
  node distance=2em, auto,
  lat/.style={draw=black, circle, minimum size=2.5em},
  det/.style={draw=black, rectangle, minimum size=2.5em},
  obs/.style={circle, draw=black, fill=black!20, minimum size=2.5em},
  gen/.style={->, -{Stealth[length=.5em, inset=0pt]}},
  inf/.style={dashed, ->, -{Stealth[length=.5em, inset=0pt]}},
]
\node[obs, inner sep=.02em] (r1) {$r_1$};
\node[obs, right=of r1, inner sep=.02em] (r2) {$r_2$};
\node[lat, right=of r2, inner sep=.02em] (r3) {$r_3$};
\node[obs, above=of r1] (o1) {$o_1$};
\node[obs, above=of r2] (o2) {$o_2$};
\node[lat, above=of r3] (o3) {$o_3$};
\node[lat, above=of o1] (x1) {$s^+_1$};
\node[lat, above=of o2] (x2) {$s^+_2$};
\node[lat, above=of o3] (x3) {$s^+_3$};
\node[lat, above=of x1] (s1) {$s^-_1$};
\node[lat, above=of x2] (s2) {$s^-_2$};
\node[lat, above=of x3] (s3) {$s^-_3$};
\node[obs, above=of s1] (a1) {$a_1$};
\node[lat, above=of s2] (a2) {$a_2$};
\path (x1) edge[gen] node {} (o1);
\path (s1) edge[gen, bend right] node {} (o1);
\path (x1) edge[gen, bend right] node {} (r1);
\path (x2) edge[gen] node {} (o2);
\path (s2) edge[gen, bend right] node {} (o2);
\path (x2) edge[gen, bend right] node {} (r2);
\path (x3) edge[gen] node {} (o3);
\path (s3) edge[gen, bend right] node {} (o3);
\path (x3) edge[gen, bend right] node {} (r3);
\path (x1) edge[gen] node {} (x2);
\path (x2) edge[gen] node {} (x3);
\path (s1) edge[gen] node {} (s2);
\path (s2) edge[gen] node {} (s3);
\path (a1) edge[gen] node {} (x2);
\path (a2) edge[gen] node {} (x3);
\path (a1) edge[gen] node {} (s2);
\path (a2) edge[gen] node {} (s3);
\path (s1) edge[red!15, gen, bend right=50, line width=2, dashed] node{} (r1);
\end{tikzpicture}}
\caption{Task-Informed MDP}\label{fig:ti_mdp}
\end{subfigure}%
\caption{
(a)~The graphical model of an MDP.
(b)~Task-Informed MDP (TiMDP). The state space decomposes into two components: \(s^+_t\) captures the task-relevant features, whereas \(s^-_t\) captures the task-irrelevant features. The cross-terms between \(s^{\small+/-}\) are removed by imposing a factored MDP assumption. The red arrow indicates an adversarial loss to discourage \(s^-\) from picking up reward relevant information.}
\label{fig:models}
\end{figure}

\paragraph{Task Informed MDP}
In many real-world problems, the state space of MDP cannot be directly accessed but needs to be inferred from high-dimensional sensory observations. Figure~\ref{fig:mdp} shows the graphical model describing this common scenario. To explicitly segregate task-relevant and irrelevant factors, we propose to model the latent embedding space $\mathcal{S}$ with two components: a task-relevant component $\mathcal{S}^+$ and a task-irrelevant component $\mathcal{S}^-$. We assume that the reward is fully determined by the task-relevant component \(r: \mathcal S^+ \mapsto \mathbb R\), and the task-irrelevant component contains no information about the reward: $\mathrm{MI}(r_t; s^{-}_{t}) = 0, \forall t$. 

In the most general case, $s_{t+1}^-$ can depend on $s_{t}^+$ and $s_{t+1}^+$ can depend on $s_t^-$. However, in many realistic scenarios the task-relevant and distractor features evolve independently (e.g. the cars on the road vs leaves flickering in the wind) and thus follow factored dynamics~\citep{guestrin2003efficient,pitis2020counterfactual}. Such a situation greatly simplifies model learning. For this reason we further incorporate this factored structure into our formulation through the assumption: $p(s_{t+1}|s_{t}, a_{t}) = p(s^+_{t+1}|s^+_t, a_{t}) p(s^-_{t+1}|s^-_t, a_{t})$.

The resulting MDP, which we call \textbf{Task Informed MDP} (TiMDP) is illustrated in Figure~\ref{fig:ti_mdp}. Note that both \(\mathcal S^+\) and \(\mathcal S^-\) generate the observation \(\mathcal O\), and both forward models \(p(s^+_{t+1}\vert s^+_t, a_t)\) and \(p(s^-_{t+1}\vert s^-_t, a_t)\) admit the agent's actions. For clarity, we summarize the assumptions we have made into \cref{table:timdp_assumptions}. 

\begin{table}[t]
\label{table:timdp_assumptions}
\caption{Assumptions for TiMDP}%
\vspace{-0.5em}%
\begin{tabular}{l|l}
	\toprule[0.6pt]
	\textbf{TiMDP} & \textbf{Details (dynamics)}\\
	\midrule[0.65pt]
	\renewcommand{\arraystretch}{1.6}
 $\scriptsize o_t = f(s^{+}_{t}, s^{-}_{t})$ & Both \(\mathcal S^{\small -/+}\) contribute to \(\mathcal O\) \\
 \midrule
 \(r: \mathcal S^+ \mapsto \mathbb R\) & \(r\) only depends on \(\mathcal S^+\) \\
 \midrule
 $\mathrm{MI}(r_t; s^{-}_{t}) = 0$ & \(\mathcal S^-\) does not inform the task \\
\midrule
 $\scriptsize\begin{array}{r@{}l@{}}%
p(s_{t+1}|&{}s_{t}, a_{t}) = \\
                &{}p(s^+_{t+1}|s^+_t, a_{t}) \\
                &{}p(s^-_{t+1}|s^-_t, a_{t})
  \end{array}$ & \(s^{\small +/-}_{t+1}\) has no dependency on \(s^{\small -/+}_t\) \\
\bottomrule[0.6pt]
\end{tabular}%
\renewcommand{\arraystretch}{1}
\end{table}
\begin{figure*}
\centering
\subcaptionbox{%
Learning \textbf{Task Informed World Models}\label{fig:world_model_learning}%
}{%
    \includegraphics[scale=0.22]{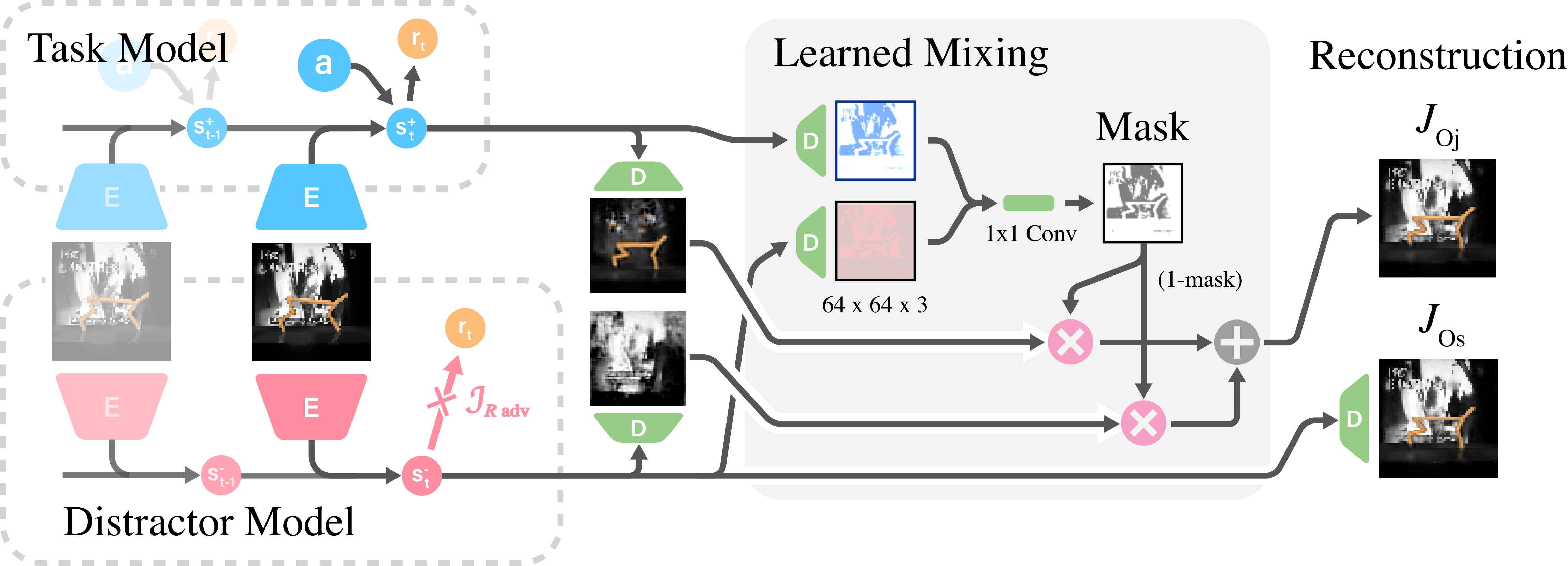}
}\hfill
\subcaptionbox{%
    \textbf{Policy Learning} \quad only unrolls in \(\mathcal S^+\).\label{fig:policy_learning}
}{%
    \includegraphics[scale=0.22]{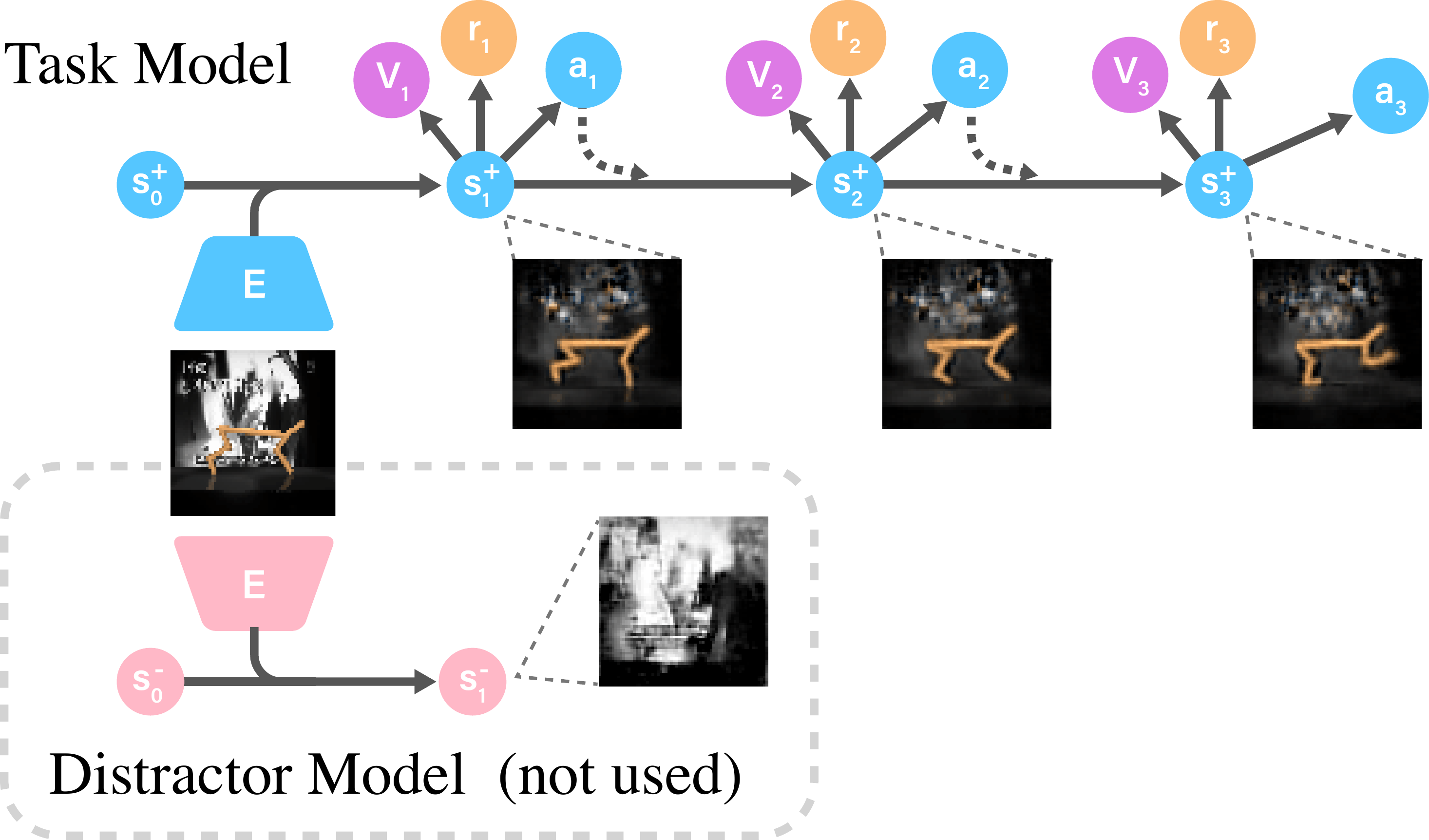}
}
\caption{Components of Task Informed Abstraction Learning. (a) From the dataset of past experience, TIA uses the reward to factor the MDP into a task-relevant world model and a task-irrelevant one. (b) Only the forward dynamics in \(s^+_{t}\) is used during policy learning. The Policy is trained using back-propagation through time. Note that the images are shown just for demonstration purposes and are not generated during policy learning.}
\end{figure*}

\begin{equation}
\small
\begin{aligned}
&\mathcal{J}_{\mathrm{T}}
  \defined \mathbb{E}_p({\sum_t\Big(\mathcal{J}_{\mathrm{Oj}}^t+\mathcal{J}_{\mathrm{R}}^t+\mathcal{J}_{\mathrm{D}}^t\Big)}) \\ 
&\mathcal{J}_{\mathrm{S}}
  \defined \mathbb{E}_p({\sum_t\Big(\mathcal{J}_{\mathrm{Oj}}^t + \mathcal{J}_{\mathrm{Os}}^t + \mathcal{J}_{\mathrm{Radv}}^t+\mathcal{J}_{\mathrm{Ds}}^t\Big)}) \\ 
& \mathcal{J}_{\mathrm{Oj}}^t
  \defined\ \ln\q(o_t|s^{+}_{t}, s^{-}_{t}) \qquad
  \mathcal{J}_{\mathrm{Os}}^t
  \defined\ \lambda_{Os} \ln\q(o_t|s^{-}_{t}) \\
& \mathcal{J}_{\mathrm{R}}^t
  \defined\ \ln\q(r_t|s^{+}_{t}) \qquad
  \mathcal{J}_{\mathrm{Radv}}^t
  \defined\ - \lambda_{\mathrm{Radv}} \max_q\ln\q(r_t|s^{-}_{t}) \\
& \mathcal{J}_{\mathrm{D}}^t
  \defined\ -\beta\KL{\p(s^{+}_{t}|s^{+}_{t-1},a_{t-1},o_t)}{\q(s^{+}_{t}|s^{+}_{t-1},a_{t-1})} \\
& \mathcal{J}_{\mathrm{Ds}}^t
  \defined\ -\beta\KL{\p(s^{-}_{t}|s^{-}_{t-1},a_{t-1},o_t)}{\q(s^{-}_{t}|s^{-}_{t-1},a_{t-1})} \\
\label{eq:disen_wm_loss}
\end{aligned}
\end{equation}%
Our method involves learning two models: one model captures the task-relevant state component $s^{+}_{t}$, which we call the \textit{task model}. The other model captures the task-irrelevant state component $s^{-}_{t}$, which we call the \textit{distractor model}. The learning objective for these two models are denoted by $\mathcal{J}_{\mathrm{T}}$ and $\mathcal{J}_{\mathrm{S}}$ (task and distractor), and expanded in \cref{eq:disen_wm_loss}. A visual illustration is provided in~\cref{fig:world_model_learning}. We will explain each component in the following section.

\textbf{Reward Dissociation} for the distractor model is accomplished via the adversarial objective $\mathcal{J}_{\mathrm{Radv}}^t$. This is a minimax setup where we interleave optimizing the distractor model's reward prediction head (for multiple iterations/training step) with the training of the distractor model. While the reward prediction head is trained to minimize the reward prediction loss $-\ln q(r_t|s_t^-)$, the distractor model maximizes this objective so as to exclude reward-correlated information from its learned features~\citep{ganin2015unsupervised}. The reward prediction loss is computed using $\ln \mathcal{N}(r_t;\hat{r}_t, 1)$, where $\mathcal{N}(\cdot;\mu, \sigma^2)$ denotes the Gaussian likelihood, and $\hat{r}_t$ is the predicted reward. 

\paragraph{Cooperative Reconstruction} By jointly reconstructing the image, the distractor model that's biased towards capturing task-irrelevant information will enable the task model to focus on task-relevant features. We implement joint reconstruction through the objective $\mathcal{J}_{\mathrm{Oj}}^t$. Starting with a sequence of observation and actions \(\{o_{[<t]}, a_{[<t]}\}\), we first pass this sequence through the two separate recurrent state space model (RSSM, \citealt{hafner2019learning}) to produce the states \(s_t^+\)and \(s_t^-\), which are then used to decode two images $\hat{o}_t^+$ and $\hat{o}_t^-$ given the observation $o_t$. The joint reconstruction is achieved through a learned mixing with a mask $M_t$, which we found to be simple and effective.

We produce the mask $M_t$ by letting the two image decoders additionally produce two tensors of size $H \!\times\! W \!\times\! 3$, where $H, W$ are the height and width of the observed image. These two tensors are concatenated channel-wise to obtain a tensor of size $H \!\times\! W \!\times\! 6$, then is passed through a $1\!\times\!1$ convolution layer followed by sigmoid activation to obtain the \(H\!\times\!W\!\times\!1\) mask $M_t$ with value between $(0,1)$. The final reconstruction is obtained through $\hat{o}_t = \hat{o}_t^+ \odot M_t + \hat{o}_t^- \odot (1-M_t)$, where $\odot$ denotes element-wise product. The reconstruction objective is computed as $\ln \mathcal{N}(o_t;\hat{o}_t, 1)$. An illustration of the mixing process is in \cref{fig:world_model_learning}.

\paragraph{Distractor-model-only Reconstruction} One failure mode of the formulation being discussed is a degenerate solution where a distractor model that captures no information at all can still satisfy the two objectives described above. This would result in the task model reconstructing the entire observation by itself. To avoid such degeneracy, we add an additional image decoder that encourages the distractor model to reconstruct the entire input observation by itself, so as to capture as much information about the observation as possible. This reconstruction loss is denoted as \(\mathcal J_{Os}^t\).

\paragraph{Policy Learning} is similar to Dreamer, except that we replace the world model with the task model. This way, the forward predictions are only made in the \(\mathcal S^+\) subspace.
\begin{equation}
\begin{aligned}
\makebox[5em][l]{Action model:} && &a_\tau \sim\ \pr[q_\phi](a_\tau|s^+_\tau) \\
\makebox[5em][l]{Value model:} && &v_\psi(s^+_\tau) \approx \mathbb{E}_{q(\cdot|s^+_\tau)}({\textstyle\sum_{\tau=t}^{t+H}\gamma^{\tau-t}r_\tau})
\label{eq:dreamer_action_value_model}
\end{aligned}
\end{equation}%
An illustration of the policy learning stage is in \cref{fig:policy_learning}.
\section{Experiments}

\begin{figure*}[t]
\providecommand{\width}{}
\renewcommand{\width}{.31\textwidth}
\centering
\begin{subfigure}[t]{0.20\textwidth}
\includegraphics[height=152pt]{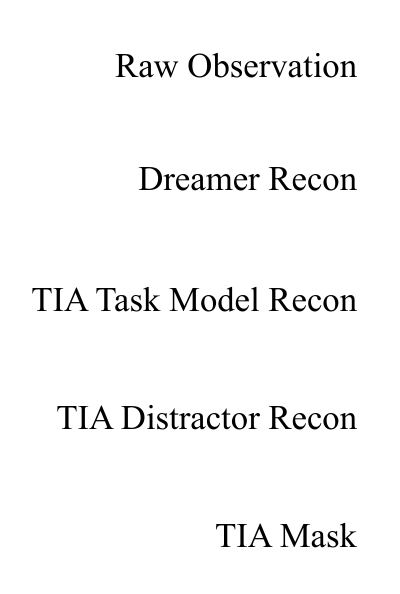}
\end{subfigure}\hfill%
\begin{subfigure}[t]{0.26\textwidth}
\includegraphics[height=152pt]{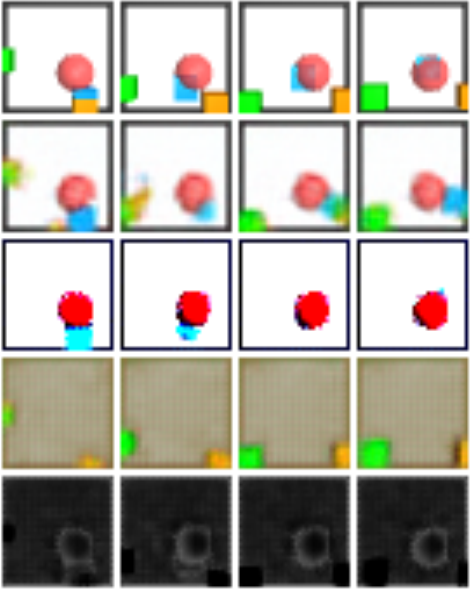}
\caption{ManyWorld}
\label{fig:rendering_manyworld}
\end{subfigure}\hfill%
\begin{subfigure}[t]{0.26\textwidth}
\includegraphics[height=152pt]{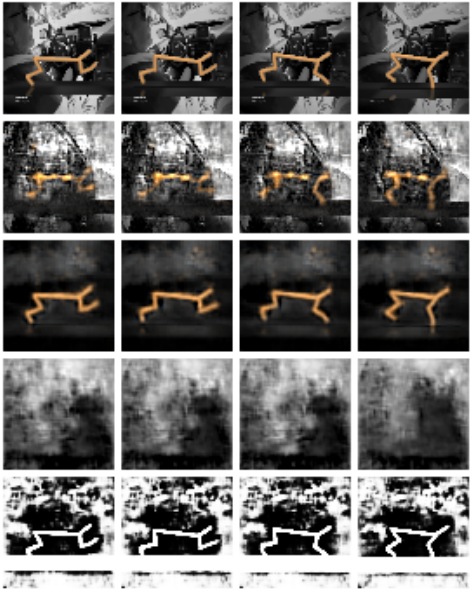}
\caption{Cheetah Run}
\label{fig:rendering_cheetah_run}
\end{subfigure}\hfill%
\begin{subfigure}[t]{0.26\textwidth}
\includegraphics[height=152pt]{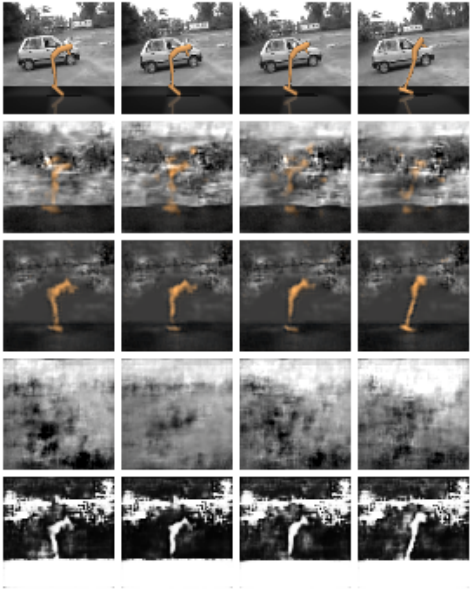}
\caption{Hopper Stand}
\label{fig:rendering_hopper_stand}
\end{subfigure}\hfill%
\caption{Visualizing the information represented by Dreamer~\cite{hafner2019dream} and the task and distractor models of our method on several environments. (a) In the ManyWorld environments, Dreamer mistakes the distractor (yellow) for the target object (blue). The task model of TIA isolates the target object (blue) and the goal (red). (b, c) Dreamer's capacity is consumed at reconstructing the irrelevant video background, and it fails to capture the agent's outline, which is the task-relevant information. In all domains, Dreamer reconstruction tries to capture every pixel of the raw observation but misses task-relevant information. TIA is able to capture task-relevant information with the task model and task-irrelevant information with the distractor model.}
\label{fig:rendering}
\vspace*{-1ex}
\end{figure*}
\begin{figure*}
    \centering
    \includegraphics[width=\textwidth]{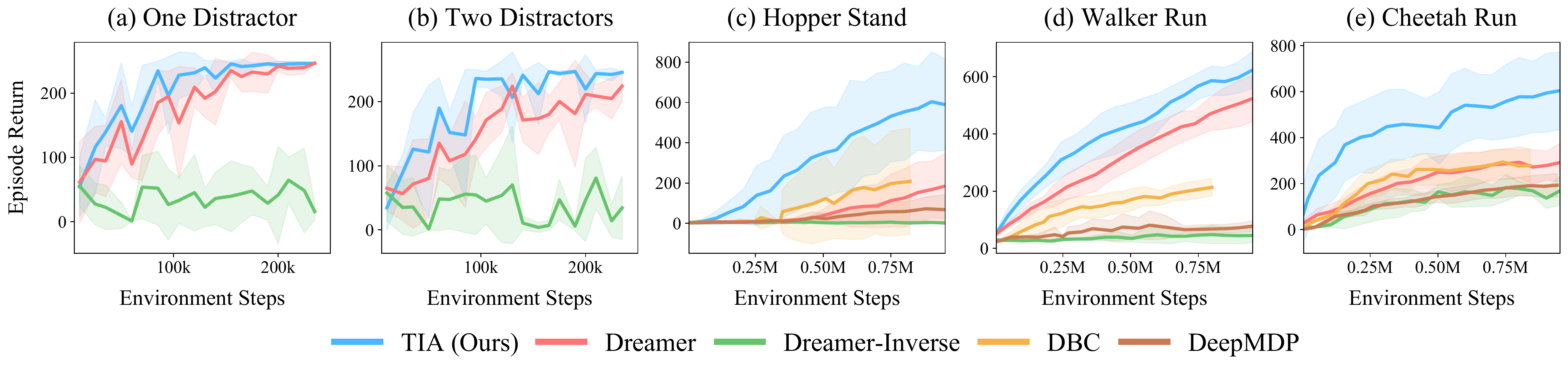}
    \caption{Our method consistently outperforms Dreamer and other baseline methods in a variety of visual control tasks with distraction. The curves show mean and standard deviation, over five seeds for TIA, Dreamer, and Dreamer-Inverse. Results for DBC and DeepMDP are adopted from results reported in \cite{zhang2020learning} and used ten seeds. Our method is effective for both ManyWorld environments (a,b), which contains confusing distracting objects that look similar to the task-relevant components; and the DMC tasks with natural video backgrounds (c,d,e), where the distracting background contains rich information that would consume significant model capacity to capture.}
    \label{fig:learning_manydmc}
\end{figure*}%

Our empirical evaluation aims to answer if the proposed method outperforms existing methods when learning in environments with visually complex backgrounds containing task-irrelevant information. For this purpose, we make use of three environments that are described in Section~\ref{sec:environments}. We compare our method against several baselines described in Section~\ref{sec:baselines}. Our code is available at \hyperlink{https://github.com/kyonofx/tia}{https://github.com/kyonofx/tia}.  

\subsection{Environments}
\label{sec:environments}
\begin{figure}[t]
    \includegraphics[width=0.23\textwidth]{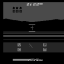}\hfill
    \includegraphics[width=0.23\textwidth]{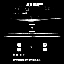}%
    \caption[Robotank Rendering]{(Left) Raw observation of Atari Robotank. (Right) The task model of TIA emphasizes task-relevant information such as the crosshair and the radar for tracking enemies, while ignoring task-irrelevant information such as textures in the raw observation.}
    \label{fig:rendering_robotank}
\end{figure}

\paragraph{ManyWorld} (\cref{fig:rendering_manyworld})
We first experimented on a custom environment suite called ManyWorld where the amount of distraction can be easily controlled. The task is to move a target block (in {\color{blue}blue}) to a goal location. The goal is visually indicated by a translucent {\color{red}red} sphere. Other objects are distractors. The amount of distraction can be controlled by varying the number and / or the dynamics of these objects. We turned off physical collisions between objects, but they visually occlude each other if they are present at overlapping positions. The task-relevant objects and the distractor objects are visually similar, necessitating additional effort to resolve when learning a world model. The observation is RGB image of size $32 \times 32 \times 3$.

\paragraph{Kinematic Control with Natural Video Distraction} (\cref{fig:rendering_cheetah_run}) We consider the DeepMind Control (DMC) suite with natural video background from the Kinetics dataset \cite{kay2017kinetics} used in prior work \cite{zhang2020learning}, which was introduced specifically to test learning under natural visual distraction. These control tasks involve different types of challenges, such as long planning horizon (Hopper), contact and collision (Walker), and larger state/action space (Cheetah). The natural video backgrounds in this test suite contains many factors of variation which makes world-model learning challenging and negatively affects downstream policy optimization. The observation is RGB image of size $64 \times 64 \times 3$. The video backgrounds are from the class ``driving car'' and are grayscale as in \citet{zhang2020learning}.

\paragraph{Arcade Learning Environments (ALE)} or Atari, (\cref{fig:rendering_robotank}) is a standard benchmark for vision-based control. The visuals of these games naturally contain many distractor objects that are irrelevant to the game objective. Our limited compute resources only allowed us to experiment on six games. Each seed takes ten days on a V100 Volta GPU. We report results on games where state-of-the-art model-based algorithms perform significantly worse than model-free algorithms or human performance in the hope of closing this gap. The observation is a grayscale image of size $64 \times 64 \times 1$ as in \citet{hafner2020mastering}.

\subsection{Baseline Methods}
\label{sec:baselines}
We compare the proposed method against both model-based and model-free baselines. Some of the baseline methods were proposed specifically to tackle learning in the presence of visual distractions. In particular we compare against \textbf{Dreamer} \cite{hafner2019dream} which is a state of the art model-based algorithm on DMC. On ALE, we compare against an improved variant \textbf{Dreamer(V2)}~\citep{hafner2020mastering}. We compare against a strong model-free method, \textbf{Deep Bisimulation for Control (DBC)} \cite{zhang2020learning}, which uses the bisimulation metric developed specifically to be invariant to task-irrelevant features. Finally, we also include \textbf{DeepMDP} \cite{gelada2019deepmdp} which augments the representation with a forward model, then uses model-free, distributional Q learning for the policy. The DeepMDP and DBC results are adapted from \citet{zhang2020learning}. 

\textbf{Representation learning through an inverse model} 
Inverse model take the observations $o_t$ and $o_{t+1}$ as input and predict the intervening action $a_t$ \cite{agrawal2015learning, jayaraman2015learning}. To investigate if features learned by inverse model suffice for learning task-relevant features, we constructed the \textit{Dreamer-Inverse} model. In this model, the learning objective becomes the following: \(\mathcal{J}^t_\text{Inverse} = \mathcal J^t_{\mathrm{inv}} + \mathcal J_R^t + \beta \mathcal J^t_D\) where \(  \mathcal{J}_{\mathrm{inv}}^t \defined\ \ln\q(a_t|s_t, s_{t+1})\) is the inverse model objective and
\(\mathcal{J}_{\mathrm{D}}^t\) is the dynamics regularizer described in Section~\ref{sec:preliminaries}.

\subsection{Can TIA disassociate task-irrelevant information?}
\label{subsec:policy_learning_curves}
We first evaluate our method on the ManyWorld domain, where relevant information comprises the agent (blue block) and the goal (red sphere). \cref{fig:rendering_manyworld} provides a qualitative comparison of the information represented by TIA and the baseline method of \textit{Dreamer}. 
In many cases, \textit{Dreamer} mistakes the distractor (yellow) for the agent (blue). On the other hand, the task model of TIA isolates the agent (blue) and the goal (red), and ignores the rest. On the other hand, the TIA's distractor model successfully captures the distractors (yellow and green). The successful dissociation of task-irrelevant information is confirmed by quantitative results in \cref{fig:learning_manydmc} (a) and (b) showing that TIA outperforms the baselines, and the performance gap increases with the number of distractors. 

Next, we considered the DMC domains with natural video distractions. We report the learning curves of \textit{Cheetah Run}, \textit{Walker Run} and \textit{Hopper Stand} are in \cref{fig:learning_manydmc}. The image reconstruction results \cref{fig:rendering_cheetah_run} and \cref{fig:rendering_hopper_stand} show that Dreamer performs poorly in capturing the full state of the agents and is distracted by the background. In contrast, the task model of our TIA method accurately recovers the relevant part of the raw observation, which happens to be the agent's body in these examples. Quantitative performance reported in \cref{fig:learning_manydmc} (c,d,e) clearly shows that TIA outperforms strong baseline methods described in Section~\ref{sec:baselines}. Overall, these results suggest that TIA is the new state-of-the-art in learning from cluttered observations. Results on additional DMC environments are included in the appendix.

\subsection{Results on Atari}
\label{sec:atari_results}
\begin{figure*}[t!]
    \centering
    \includegraphics[width=\textwidth]{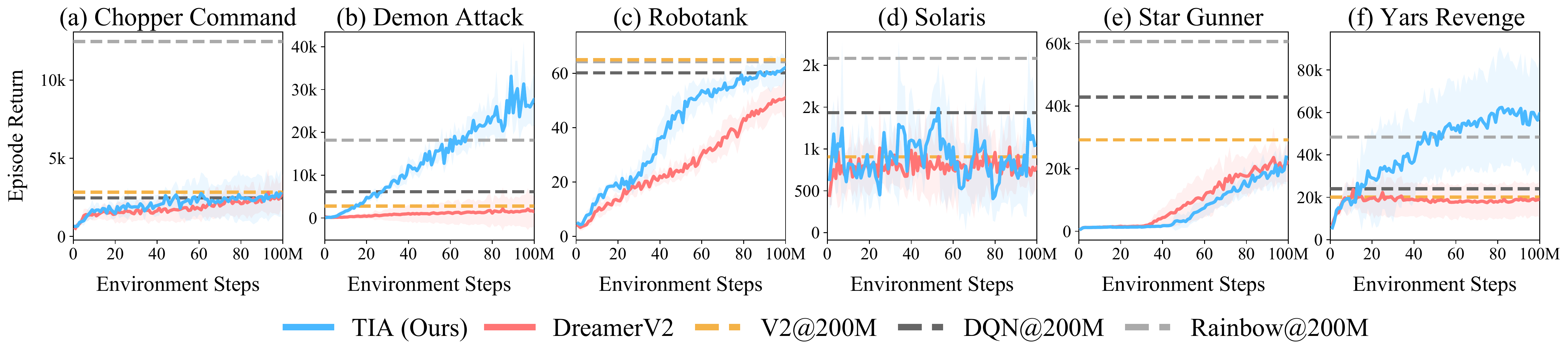}
    \caption{Atari performance at 100M steps. TIA significantly improves policy learning in Demon Attack, Robotank, and Yars Revenge, in which DreamerV2 fails to learn or has inferior sample complexity. We also add the performance of DreamerV2 at 200M steps, along with the performance of two model-free algorithms, DQN and Rainbow DQN, at 200M steps.}
    \label{fig:atari}
\end{figure*}%

In environments tested so far the irrelevant factors were manually injected. Due to human intervention, it is possible these tasks were biased, which our method exploited. To investigate performance in more natural settings, we experimented on Atari games that innately contain significant visual distraction. E.g., images of \textit{Robotank} game contain several visual signatures that change during the game: the number of enemy tanks destroyed, the rotating radar scanner, the green sprites, and so on. We evaluated performance on six Atari games that are known to be challenging for model-based methods~\cite{hafner2019dream} without any hyperparameter tuning (i.e., just a single value chosen based on the intuition described in Section~\ref{sec:hyper_sensitivity}). 

The results reported in \cref{fig:atari} demonstrate that we substantially outperform the strong baseline of DreamerV2. Furthermore, for the games of \textit{Demon Attack}, \textit{Robotank} and \textit{Yars Revenge}, we match the performance or outperform strong model-free baselines of DQN/Rainbow \cite{mnih2013playing, hessel2018rainbow} trained for 200M steps, while our method is only trained for 100M steps. These results convincingly demonstrate the superiority of our method in visually cluttered domains. \cref{fig:rendering_robotank} shows the image reconstruction of TIA's task model in \textit{Robotank}.

\subsection{Hyperparameter Selection}
\label{sec:hyper_sensitivity}
The two important hyperparameters are \(\lambda_\mathrm{Radv}\) and \(\lambda_\mathrm{Os}\). The balance of these two terms enables the distractor model to capture the task-irrelevant part of states. One particular mode of failure is when the distractor model takes over the reconstruction. It strips the task model from task-relevant information, thus preventing the policy from learning meaningful behavior. Our reward dissociation scheme relies on informative reward signals to work. Yet, at the beginning of training, the reward collected by a random policy tend to be sparse and noisy, making $\lambda_\mathrm{Radv}$ less effective at preventing distractor model from dominating. This scenario suggests using a large $\lambda_\mathrm{Radv}$ at the beginning of training and slowly increasing the weight \(\lambda_\mathrm{Os}\) for the distractor reconstruction loss.

\begin{figure}[t]
    \centering
    \includegraphics[width=0.48\textwidth]{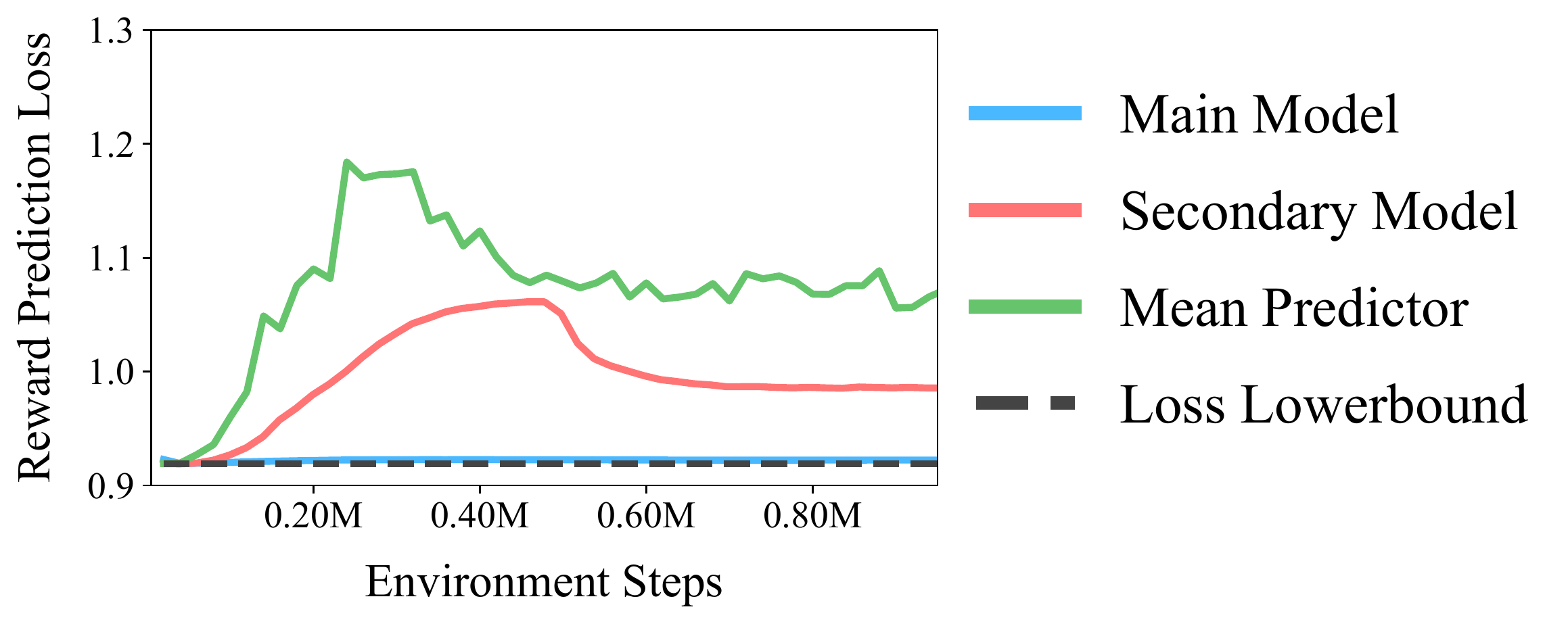}%
    \caption{\textbf{Reward Dissociation During Learning}\;\; 
    We plot the negative log-likelihood loss of the reward prediction $-\ln p(r)$, (lower bounded by $0.92$) of a mean predictor, the reward prediction module of the task model, and the reward prediction module of the distractor model. The features from the task model contain sufficient information for reward prediction. The performance of the reward predictor for the distractor model follows the same trend as the mean predictor, indicating that the features learned by the distractor model are reward-independent.}
    \label{fig:reward_dissociation}
\end{figure}

On the other extreme the distractor can become degenerate and capture no information of the input observations. In this case, TIA is equivalent to Dreamer. To avoid such degeneracy we would increase $\lambda_\mathrm{Os}$ so that the distractor model is encouraged to capture more information.

To gain more insight into whether the reward dissociation process works as expected, we computed the how much reward specific information is captured by the distractor model capture during. For this we recorded the reward prediction error measured using the log-likelihood (as unit Gaussian) in \cref{fig:reward_dissociation}. Since the error is unit-less, for the sake of comparison, we estimated the upper bound of the prediction error using the running average of the reward in all historical data. This corresponds to an uninformed reward predictor (the ``Mean Predictor'') that always guesses the average. The reward prediction loss of the distractor model trails behind this upper bound during learning~(see~\cref{fig:reward_dissociation}, in {\color{red}red}), whereas the reward prediction loss of the main model remains small (in {\color{blue}blue}), slightly above the loss lower bound, which equals to \(-\ln \mathcal N (0;\; 0, 1) = -\ln{\frac 1 {\sqrt{2\pi}}}\approx 0.92.\) This result indicates that the task model learns task-relevant information, while the distractor model's learned feature is indeed reward-independent. 

\section{Other Related Works}

State-of-the-art deep reinforcement learning algorithms often jointly optimize the discounted return together with an auxiliary representation learning objective such as image reconstruction~\citep{watter2015embed, wahlstrom2015pixels} or contrastive learning (\citealt{sermanet2018time, oord2018representation, yan2020learning, lee2020predictive, mazoure2020deep}). In model-based reinforcement learning methods, reward and value prediction \cite{oh2017value, racaniere2017imagination, silver2016predictron, tamar2016value, feinberg2018model, schrittwieser2019mastering} are also shown to improve performance, along with learning a world model. More recently, data augmentations are found to improve sample complexity \cite{laskin2020reinforcement, kostrikov2020image, raileanu2020automatic, srinivas2020curl, stooke2020decoupling} by taking advantage of domain knowledge of symmetry transformations in the space of data. A recent benchmark \cite{stone2021distracting} however, shows that data augmentation helps but still performs poorly in the presence of complex visual distractions. We do find data augmentation to be an orthogonal and complementary approach to our proposal. Our approach focuses on finding ways to inform representation learning for RL which features are more useful for the task, and therefore should be learned first.

Block MDPs\cite{du2019provably}, bisimulation \cite{givan2003equivalence, dean1997model} and bisimulation metrics \cite{ferns2004metrics, ferns2011bisimulation, ferns2014bisimulation} exploit additional structure to learn state abstractions, leading to impressive empirical gains on natural scene domains\cite{zhang2020learning, gelada2019deepmdp, agarwal2021contrastive}. The work presented here falls under the broad umbrella of learning state and temporal abstractions and is the most closely related to \textit{utile distinction}~\cite{McCallum1997util-distinction} which is limited to finite-state machines. Like \textit{utile distinction}, TIA is distinct from DBC and bisimulation embeddings in that TIA relies on the task specification for feature separation. It is not concerned about generalization across a set of MDPs related by dynamics.

The spiritual nearest neighbor to our work is the idea of using a primary-bias duo for model debiasing in supervised learning ~\cite{clark2019don, cadene2019rubi, wang2019learning, he2019unlearn, bahng2020learning, clark2020learning}. These methods remove dataset bias by imposing independence constraints between the primary model and a model that's biased by design. Our approach focuses on RL and uses two models for removing task-irrelevant features for policy learning. Our implementation extends recent work in model-based RL from pixels \cite{watter2015embed, finn2017deep, banijamali2018robust, ha2018world, kaiser2019model, hafner2019dream, hafner2020mastering, srinivas2018universal, pertsch2020long}.
\section{Conclusion and Discussion}
\label{sec:conclusion}

This work demonstrates that the TiMDP formulation offers a way for a model-based agent to explain away task-irrelevant information through a distractor model, and successfully learn from cluttered visual inputs. Our approach of learning \ourmethod (TIA) outperforms previous state-of-the-art model-based RL methods on two standard benchmarks. TIA reduces to Dreamer when its distractor model degenerates and captures no information. TIA would outperform Dreamer when the hyperparameters are well-tuned and the disentangled state representation would help policy learning.

If we are able to distinguish task-relevant features from irrelevant ones, one might expect that such a scheme would also lead to better transfer to new scenarios. To investigate this, we ran experiments on DMC tasks, where the train and test backgrounds consisted of different videos. We found that when the train and test sets had different sets of natural video backgrounds, our method could easily generalize. However, when we trained on random noise background and test on natural video backgrounds without any fine-tuning, the performance suffered. This result is unsurprising, and the transfer problem can be alleviated by training on diverse data. Further, in our current setup, the learning of task-relevant information only happens during training. However, there is no mechanism to adapt what is relevant/irrelevant at test time. Investigating test-time adaptation methods can improve transfer performance to out-of-distribution situations.  

A second issue worth mentioning is that while one set of hyperparameters works well across Atari games, for the DMC environments, the choice of hyperparameters, $\lambda_\mathrm{Radv}$ and $ \lambda_\mathrm{Os}$, is domain-dependent. We discussed good practices for choosing these hyperparameters in \cref{sec:hyper_sensitivity}. We hypothesize that based the practice documented here, in the future, it might be possible to automatically tune the hyperparameters by considering the reconstruction and reward-prediction loss of the two models that constitute TIA.

It is possible that the factored dynamics assumption in TiMDP is too strong. Take ManyWorld as an example; this assumption is violated when distractor blocks interact with the target and are no longer task-independent. In practice, TIA implements a relaxed version of the factored dynamics constraint via cooperative reconstruction: Features that interact weakly (such as occlusion) can be filtered into the distractor model (see \cref{fig:rendering_manyworld}), whereas ones that interact strongly (such as via collision) are highly task-relevant, therefore will be included in the task model.

Our goal in this work was to build agents that operate from complex visual imagery. Another area of potential investigation is to characterize performance as a function of the sparsity of reward signals. We hypothesize that a low-noise reward signal is a key factor for the robustness of TIA, and the performance might drop in very sparse reward scenarios. Developing methods that can overcome this ``potential" challenge is another avenue for future work.

\section*{Acknowledgement}

We would like to thank Amy Zhang for sharing the performance data of DBC and DeepMDP and Danijar Hafner for sharing the performance data of DreamerV2, DQN, and Rainbow. This work is supported by the MIT-IBM grant on adversarial learning of multi-modal and structured data, the MIT-IBM Quest Program, the DARPA Machine Common Sense Program, and the National Science Foundation under Cooperative Agreement PHY-2019786 (The NSF AI Institute for Artificial Intelligence and Fundamental Interactions, http://iaifi.org/). The authors also acknowledge the MIT SuperCloud and the Lincoln Laboratory Supercomputing Center for providing HPC resources.
\bibliography{reference}

\begin{thebibliography}{65}
\providecommand{\natexlab}[1]{#1}
\providecommand{\url}[1]{\texttt{#1}}
\expandafter\ifx\csname urlstyle\endcsname\relax
  \providecommand{\doi}[1]{doi: #1}\else
  \providecommand{\doi}{doi: \begingroup \urlstyle{rm}\Url}\fi

\bibitem[Agarwal et~al.(2021)Agarwal, Machado, Castro, and
  Bellemare]{agarwal2021contrastive}
Agarwal, R., Machado, M.~C., Castro, P.~S., and Bellemare, M.~G.
\newblock Contrastive behavioral similarity embeddings for generalization in
  reinforcement learning.
\newblock In \emph{International Conference on Learning Representations}, 2021.

\bibitem[Agrawal et~al.(2015)Agrawal, Carreira, and Malik]{agrawal2015learning}
Agrawal, P., Carreira, J., and Malik, J.
\newblock Learning to see by moving.
\newblock In \emph{Proceedings of the IEEE international conference on computer
  vision}, pp.\  37--45, 2015.

\bibitem[Agrawal et~al.(2016)Agrawal, Nair, Abbeel, Malik, and
  Levine]{agrawal2016learning}
Agrawal, P., Nair, A.~V., Abbeel, P., Malik, J., and Levine, S.
\newblock Learning to poke by poking: Experiential learning of intuitive
  physics.
\newblock In \emph{Advances in Neural Information Processing Systems}, pp.\
  5074--5082, 2016.

\bibitem[Bahng et~al.(2020)Bahng, Chun, Yun, Choo, and Oh]{bahng2020learning}
Bahng, H., Chun, S., Yun, S., Choo, J., and Oh, S.~J.
\newblock Learning de-biased representations with biased representations.
\newblock In \emph{International Conference on Machine Learning}, pp.\
  528--539. PMLR, 2020.

\bibitem[Banijamali et~al.(2018)Banijamali, Shu, Bui, Ghodsi,
  et~al.]{banijamali2018robust}
Banijamali, E., Shu, R., Bui, H., Ghodsi, A., et~al.
\newblock Robust locally-linear controllable embedding.
\newblock In \emph{International Conference on Artificial Intelligence and
  Statistics}, pp.\  1751--1759. PMLR, 2018.

\bibitem[Bengio(2013)]{bengio2013representation}
Bengio, Y.
\newblock Deep learning of representations: Looking forward.
\newblock In Dediu, A., Mart{\'{\i}}n{-}Vide, C., Mitkov, R., and Truthe, B.
  (eds.), \emph{Statistical Language and Speech Processing - First
  International Conference, {SLSP} 2013, Tarragona, Spain, July 29-31, 2013.
  Proceedings}, volume 7978 of \emph{Lecture Notes in Computer Science}, pp.\
  1--37. Springer, 2013.
\newblock \doi{10.1007/978-3-642-39593-2\_1}.

\bibitem[Cadene et~al.(2019)Cadene, Dancette, Cord, Parikh,
  et~al.]{cadene2019rubi}
Cadene, R., Dancette, C., Cord, M., Parikh, D., et~al.
\newblock Rubi: Reducing unimodal biases for visual question answering.
\newblock In \emph{Advances in neural information processing systems}, pp.\
  841--852, 2019.

\bibitem[Chen et~al.(2020)Chen, Kornblith, Norouzi, and Hinton]{chen2020simple}
Chen, T., Kornblith, S., Norouzi, M., and Hinton, G.
\newblock A simple framework for contrastive learning of visual
  representations.
\newblock In \emph{International conference on machine learning}, pp.\
  1597--1607. PMLR, 2020.

\bibitem[Clark et~al.(2019)Clark, Yatskar, and Zettlemoyer]{clark2019don}
Clark, C., Yatskar, M., and Zettlemoyer, L.
\newblock Don't take the easy way out: Ensemble based methods for avoiding
  known dataset biases.
\newblock In Inui, K., Jiang, J., Ng, V., and Wan, X. (eds.), \emph{Proceedings
  of the 2019 Conference on Empirical Methods in Natural Language Processing
  and the 9th International Joint Conference on Natural Language Processing,
  {EMNLP-IJCNLP} 2019, Hong Kong, China, November 3-7, 2019}, pp.\  4067--4080.
  Association for Computational Linguistics, 2019.
\newblock \doi{10.18653/v1/D19-1418}.

\bibitem[Clark et~al.(2020)Clark, Yatskar, and Zettlemoyer]{clark2020learning}
Clark, C., Yatskar, M., and Zettlemoyer, L.
\newblock Learning to model and ignore dataset bias with mixed capacity
  ensembles.
\newblock In Cohn, T., He, Y., and Liu, Y. (eds.), \emph{Proceedings of the
  2020 Conference on Empirical Methods in Natural Language Processing:
  Findings, {EMNLP} 2020, Online Event, 16-20 November 2020}, pp.\  3031--3045.
  Association for Computational Linguistics, 2020.
\newblock \doi{10.18653/v1/2020.findings-emnlp.272}.

\bibitem[Dean et~al.(1997)Dean, Givan, and Leach]{dean1997model}
Dean, T.~L., Givan, R., and Leach, S.~M.
\newblock Model reduction techniques for computing approximately optimal
  solutions for markov decision processes.
\newblock In Geiger, D. and Shenoy, P.~P. (eds.), \emph{{UAI} '97: Proceedings
  of the Thirteenth Conference on Uncertainty in Artificial Intelligence, Brown
  University, Providence, Rhode Island, USA, August 1-3, 1997}, pp.\  124--131.
  Morgan Kaufmann, 1997.

\bibitem[Du et~al.(2019)Du, Krishnamurthy, Jiang, Agarwal, Dudik, and
  Langford]{du2019provably}
Du, S., Krishnamurthy, A., Jiang, N., Agarwal, A., Dudik, M., and Langford, J.
\newblock Provably efficient rl with rich observations via latent state
  decoding.
\newblock In \emph{International Conference on Machine Learning}, pp.\
  1665--1674. PMLR, 2019.

\bibitem[Feinberg et~al.(2018)Feinberg, Wan, Stoica, Jordan, Gonzalez, and
  Levine]{feinberg2018model}
Feinberg, V., Wan, A., Stoica, I., Jordan, M.~I., Gonzalez, J.~E., and Levine,
  S.
\newblock Model-based value estimation for efficient model-free reinforcement
  learning.
\newblock \emph{arXiv preprint arXiv:1803.00101}, 2018.

\bibitem[Ferns \& Precup(2014)Ferns and Precup]{ferns2014bisimulation}
Ferns, N. and Precup, D.
\newblock Bisimulation metrics are optimal value functions.
\newblock In \emph{UAI}, pp.\  210--219. Citeseer, 2014.

\bibitem[Ferns et~al.(2004)Ferns, Panangaden, and Precup]{ferns2004metrics}
Ferns, N., Panangaden, P., and Precup, D.
\newblock Metrics for finite markov decision processes.
\newblock In \emph{UAI}, volume~4, pp.\  162--169, 2004.

\bibitem[Ferns et~al.(2011)Ferns, Panangaden, and
  Precup]{ferns2011bisimulation}
Ferns, N., Panangaden, P., and Precup, D.
\newblock Bisimulation metrics for continuous markov decision processes.
\newblock \emph{SIAM Journal on Computing}, 40\penalty0 (6):\penalty0
  1662--1714, 2011.

\bibitem[Finn \& Levine(2017)Finn and Levine]{finn2017deep}
Finn, C. and Levine, S.
\newblock Deep visual foresight for planning robot motion.
\newblock In \emph{2017 IEEE International Conference on Robotics and
  Automation (ICRA)}, pp.\  2786--2793. IEEE, 2017.

\bibitem[Ganin \& Lempitsky(2015)Ganin and Lempitsky]{ganin2015unsupervised}
Ganin, Y. and Lempitsky, V.
\newblock Unsupervised domain adaptation by backpropagation.
\newblock In \emph{International conference on machine learning}, pp.\
  1180--1189. PMLR, 2015.

\bibitem[Gelada et~al.(2019)Gelada, Kumar, Buckman, Nachum, and
  Bellemare]{gelada2019deepmdp}
Gelada, C., Kumar, S., Buckman, J., Nachum, O., and Bellemare, M.~G.
\newblock Deepmdp: Learning continuous latent space models for representation
  learning.
\newblock In Chaudhuri, K. and Salakhutdinov, R. (eds.), \emph{Proceedings of
  the 36th International Conference on Machine Learning, {ICML} 2019, 9-15 June
  2019, Long Beach, California, {USA}}, volume~97 of \emph{Proceedings of
  Machine Learning Research}, pp.\  2170--2179. {PMLR}, 2019.

\bibitem[Givan et~al.(2003)Givan, Dean, and Greig]{givan2003equivalence}
Givan, R., Dean, T.~L., and Greig, M.
\newblock Equivalence notions and model minimization in markov decision
  processes.
\newblock \emph{Artif. Intell.}, 147\penalty0 (1-2):\penalty0 163--223, 2003.
\newblock \doi{10.1016/S0004-3702(02)00376-4}.

\bibitem[Gregor et~al.(2016)Gregor, Rezende, and
  Wierstra]{gregor2016variational}
Gregor, K., Rezende, D.~J., and Wierstra, D.
\newblock Variational intrinsic control.
\newblock \emph{arXiv preprint arXiv:1611.07507}, 2016.

\bibitem[Guestrin et~al.(2003)Guestrin, Koller, Parr, and
  Venkataraman]{guestrin2003efficient}
Guestrin, C., Koller, D., Parr, R., and Venkataraman, S.
\newblock Efficient solution algorithms for factored mdps.
\newblock \emph{Journal of Artificial Intelligence Research}, 19:\penalty0
  399--468, 2003.

\bibitem[Ha \& Schmidhuber(2018)Ha and Schmidhuber]{ha2018world}
Ha, D. and Schmidhuber, J.
\newblock Recurrent world models facilitate policy evolution.
\newblock In Bengio, S., Wallach, H.~M., Larochelle, H., Grauman, K.,
  Cesa{-}Bianchi, N., and Garnett, R. (eds.), \emph{Advances in Neural
  Information Processing Systems 31: Annual Conference on Neural Information
  Processing Systems 2018, NeurIPS 2018, December 3-8, 2018, Montr{\'{e}}al,
  Canada}, pp.\  2455--2467, 2018.

\bibitem[Hafner et~al.(2019)Hafner, Lillicrap, Fischer, Villegas, Ha, Lee, and
  Davidson]{hafner2019learning}
Hafner, D., Lillicrap, T., Fischer, I., Villegas, R., Ha, D., Lee, H., and
  Davidson, J.
\newblock Learning latent dynamics for planning from pixels.
\newblock In \emph{International Conference on Machine Learning}, pp.\
  2555--2565. PMLR, 2019.

\bibitem[Hafner et~al.(2020)Hafner, Lillicrap, Ba, and
  Norouzi]{hafner2019dream}
Hafner, D., Lillicrap, T.~P., Ba, J., and Norouzi, M.
\newblock Dream to control: Learning behaviors by latent imagination.
\newblock In \emph{8th International Conference on Learning Representations,
  {ICLR} 2020, Addis Ababa, Ethiopia, April 26-30, 2020}. OpenReview.net, 2020.

\bibitem[Hafner et~al.(2021)Hafner, Lillicrap, Norouzi, and
  Ba]{hafner2020mastering}
Hafner, D., Lillicrap, T.~P., Norouzi, M., and Ba, J.
\newblock Mastering atari with discrete world models.
\newblock In \emph{International Conference on Learning Representations}, 2021.

\bibitem[He et~al.(2019)He, Zha, and Wang]{he2019unlearn}
He, H., Zha, S., and Wang, H.
\newblock Unlearn dataset bias in natural language inference by fitting the
  residual.
\newblock In Cherry, C., Durrett, G., Foster, G.~F., Haffari, R., Khadivi, S.,
  Peng, N., Ren, X., and Swayamdipta, S. (eds.), \emph{Proceedings of the 2nd
  Workshop on Deep Learning Approaches for Low-Resource NLP,
  DeepLo@EMNLP-IJCNLP 2019, Hong Kong, China, November 3, 2019}, pp.\
  132--142. Association for Computational Linguistics, 2019.
\newblock \doi{10.18653/v1/D19-6115}.

\bibitem[Hessel et~al.(2018)Hessel, Modayil, Van~Hasselt, Schaul, Ostrovski,
  Dabney, Horgan, Piot, Azar, and Silver]{hessel2018rainbow}
Hessel, M., Modayil, J., Van~Hasselt, H., Schaul, T., Ostrovski, G., Dabney,
  W., Horgan, D., Piot, B., Azar, M., and Silver, D.
\newblock Rainbow: Combining improvements in deep reinforcement learning.
\newblock In \emph{Proceedings of the AAAI Conference on Artificial
  Intelligence}, volume~32, 2018.

\bibitem[Higgins et~al.(2016)Higgins, Matthey, Pal, Burgess, Glorot, Botvinick,
  Mohamed, and Lerchner]{higgins2016beta}
Higgins, I., Matthey, L., Pal, A., Burgess, C., Glorot, X., Botvinick, M.,
  Mohamed, S., and Lerchner, A.
\newblock beta-vae: Learning basic visual concepts with a constrained
  variational framework.
\newblock 2016.

\bibitem[Higgins et~al.(2017)Higgins, Pal, Rusu, Matthey, Burgess, Pritzel,
  Botvinick, Blundell, and Lerchner]{higgins2017darla}
Higgins, I., Pal, A., Rusu, A., Matthey, L., Burgess, C., Pritzel, A.,
  Botvinick, M., Blundell, C., and Lerchner, A.
\newblock Darla: Improving zero-shot transfer in reinforcement learning.
\newblock In \emph{International Conference on Machine Learning}, pp.\
  1480--1490. PMLR, 2017.

\bibitem[Jayaraman \& Grauman(2015)Jayaraman and
  Grauman]{jayaraman2015learning}
Jayaraman, D. and Grauman, K.
\newblock Learning image representations tied to ego-motion.
\newblock In \emph{Proceedings of the IEEE International Conference on Computer
  Vision}, pp.\  1413--1421, 2015.

\bibitem[Kaiser et~al.(2020)Kaiser, Babaeizadeh, Milos, Osinski, Campbell,
  Czechowski, Erhan, Finn, Kozakowski, Levine, Mohiuddin, Sepassi, Tucker, and
  Michalewski]{kaiser2019model}
Kaiser, L., Babaeizadeh, M., Milos, P., Osinski, B., Campbell, R.~H.,
  Czechowski, K., Erhan, D., Finn, C., Kozakowski, P., Levine, S., Mohiuddin,
  A., Sepassi, R., Tucker, G., and Michalewski, H.
\newblock Model based reinforcement learning for atari.
\newblock In \emph{8th International Conference on Learning Representations,
  {ICLR} 2020, Addis Ababa, Ethiopia, April 26-30, 2020}. OpenReview.net, 2020.

\bibitem[Kay et~al.(2017)Kay, Carreira, Simonyan, Zhang, Hillier,
  Vijayanarasimhan, Viola, Green, Back, Natsev, et~al.]{kay2017kinetics}
Kay, W., Carreira, J., Simonyan, K., Zhang, B., Hillier, C., Vijayanarasimhan,
  S., Viola, F., Green, T., Back, T., Natsev, P., et~al.
\newblock The kinetics human action video dataset.
\newblock \emph{arXiv preprint arXiv:1705.06950}, 2017.

\bibitem[Kingma \& Welling(2014)Kingma and Welling]{kingma2014stochastic}
Kingma, D.~P. and Welling, M.
\newblock Stochastic gradient vb and the variational auto-encoder.
\newblock In \emph{Second International Conference on Learning Representations,
  ICLR}, volume~19, 2014.

\bibitem[Kingma et~al.(2014)Kingma, Mohamed, Jimenez~Rezende, and
  Welling]{kingma2014semi}
Kingma, D.~P., Mohamed, S., Jimenez~Rezende, D., and Welling, M.
\newblock Semi-supervised learning with deep generative models.
\newblock In Ghahramani, Z., Welling, M., Cortes, C., Lawrence, N.~D., and
  Weinberger, K.~Q. (eds.), \emph{Advances in Neural Information Processing
  Systems 27}, pp.\  3581--3589. Curran Associates, Inc., 2014.

\bibitem[Klyubin et~al.(2005)Klyubin, Polani, and
  Nehaniv]{klyubin2005empowerment}
Klyubin, A.~S., Polani, D., and Nehaniv, C.~L.
\newblock Empowerment: A universal agent-centric measure of control.
\newblock In \emph{2005 IEEE Congress on Evolutionary Computation}, volume~1,
  pp.\  128--135. IEEE, 2005.

\bibitem[Laskin et~al.(2020{\natexlab{a}})Laskin, Lee, Stooke, Pinto, Abbeel,
  and Srinivas]{laskin2020reinforcement}
Laskin, M., Lee, K., Stooke, A., Pinto, L., Abbeel, P., and Srinivas, A.
\newblock Reinforcement learning with augmented data.
\newblock In Larochelle, H., Ranzato, M., Hadsell, R., Balcan, M., and Lin, H.
  (eds.), \emph{Advances in Neural Information Processing Systems 33: Annual
  Conference on Neural Information Processing Systems 2020, NeurIPS 2020,
  December 6-12, 2020, virtual}, 2020{\natexlab{a}}.

\bibitem[Laskin et~al.(2020{\natexlab{b}})Laskin, Srinivas, and
  Abbeel]{srinivas2020curl}
Laskin, M., Srinivas, A., and Abbeel, P.
\newblock {CURL:} contrastive unsupervised representations for reinforcement
  learning.
\newblock In \emph{Proceedings of the 37th International Conference on Machine
  Learning, {ICML} 2020, 13-18 July 2020, Virtual Event}, volume 119 of
  \emph{Proceedings of Machine Learning Research}, pp.\  5639--5650. {PMLR},
  2020{\natexlab{b}}.

\bibitem[Lee et~al.(2020)Lee, Fischer, Liu, Guo, Lee, Canny, and
  Guadarrama]{lee2020predictive}
Lee, K., Fischer, I., Liu, A., Guo, Y., Lee, H., Canny, J., and Guadarrama, S.
\newblock Predictive information accelerates learning in {RL}.
\newblock In Larochelle, H., Ranzato, M., Hadsell, R., Balcan, M., and Lin, H.
  (eds.), \emph{Advances in Neural Information Processing Systems 33: Annual
  Conference on Neural Information Processing Systems 2020, NeurIPS 2020,
  December 6-12, 2020, virtual}, 2020.

\bibitem[Mazoure et~al.(2020)Mazoure, des Combes, Doan, Bachman, and
  Hjelm]{mazoure2020deep}
Mazoure, B., des Combes, R.~T., Doan, T., Bachman, P., and Hjelm, R.~D.
\newblock Deep reinforcement and infomax learning.
\newblock In Larochelle, H., Ranzato, M., Hadsell, R., Balcan, M., and Lin, H.
  (eds.), \emph{Advances in Neural Information Processing Systems 33: Annual
  Conference on Neural Information Processing Systems 2020, NeurIPS 2020,
  December 6-12, 2020, virtual}, 2020.

\bibitem[McCallum(1997)]{McCallum1997util-distinction}
McCallum, R.
\newblock Reinforcement learning with selective perception and hidden state.
\newblock 1997.

\bibitem[Mnih et~al.(2013)Mnih, Kavukcuoglu, Silver, Graves, Antonoglou,
  Wierstra, and Riedmiller]{mnih2013playing}
Mnih, V., Kavukcuoglu, K., Silver, D., Graves, A., Antonoglou, I., Wierstra,
  D., and Riedmiller, M.
\newblock Playing atari with deep reinforcement learning.
\newblock \emph{arXiv preprint arXiv:1312.5602}, 2013.

\bibitem[Oh et~al.(2017)Oh, Singh, and Lee]{oh2017value}
Oh, J., Singh, S., and Lee, H.
\newblock Value prediction network.
\newblock In Guyon, I., von Luxburg, U., Bengio, S., Wallach, H.~M., Fergus,
  R., Vishwanathan, S. V.~N., and Garnett, R. (eds.), \emph{Advances in Neural
  Information Processing Systems 30: Annual Conference on Neural Information
  Processing Systems 2017, December 4-9, 2017, Long Beach, CA, {USA}}, pp.\
  6118--6128, 2017.

\bibitem[Oord et~al.(2018)Oord, Li, and Vinyals]{oord2018representation}
Oord, A. v.~d., Li, Y., and Vinyals, O.
\newblock Representation learning with contrastive predictive coding.
\newblock \emph{arXiv preprint arXiv:1807.03748}, 2018.

\bibitem[Pathak et~al.(2017)Pathak, Agrawal, Efros, and
  Darrell]{pathak2017curiosity}
Pathak, D., Agrawal, P., Efros, A.~A., and Darrell, T.
\newblock Curiosity-driven exploration by self-supervised prediction.
\newblock In \emph{Proceedings of the 34th International Conference on Machine
  Learning}, pp.\  2778--2787, 2017.

\bibitem[Pertsch et~al.(2020)Pertsch, Rybkin, Ebert, Zhou, Jayaraman, Finn, and
  Levine]{pertsch2020long}
Pertsch, K., Rybkin, O., Ebert, F., Zhou, S., Jayaraman, D., Finn, C., and
  Levine, S.
\newblock Long-horizon visual planning with goal-conditioned hierarchical
  predictors.
\newblock \emph{Advances in Neural Information Processing Systems}, 33, 2020.

\bibitem[Pitis et~al.(2020)Pitis, Creager, and Garg]{pitis2020counterfactual}
Pitis, S., Creager, E., and Garg, A.
\newblock Counterfactual data augmentation using locally factored dynamics.
\newblock In Larochelle, H., Ranzato, M., Hadsell, R., Balcan, M., and Lin, H.
  (eds.), \emph{Advances in Neural Information Processing Systems 33: Annual
  Conference on Neural Information Processing Systems 2020, NeurIPS 2020,
  December 6-12, 2020, virtual}, 2020.

\bibitem[Racani{\`e}re et~al.(2017)Racani{\`e}re, Weber, Reichert, Buesing,
  Guez, Rezende, Badia, Vinyals, Heess, Li, et~al.]{racaniere2017imagination}
Racani{\`e}re, S., Weber, T., Reichert, D.~P., Buesing, L., Guez, A., Rezende,
  D., Badia, A.~P., Vinyals, O., Heess, N., Li, Y., et~al.
\newblock Imagination-augmented agents for deep reinforcement learning.
\newblock In \emph{Proceedings of the 31st International Conference on Neural
  Information Processing Systems}, pp.\  5694--5705, 2017.

\bibitem[Raileanu et~al.(2020)Raileanu, Goldstein, Yarats, Kostrikov, and
  Fergus]{raileanu2020automatic}
Raileanu, R., Goldstein, M., Yarats, D., Kostrikov, I., and Fergus, R.
\newblock Automatic data augmentation for generalization in deep reinforcement
  learning.
\newblock \emph{arXiv preprint arXiv:2006.12862}, 2020.

\bibitem[Schrittwieser et~al.(2020)Schrittwieser, Antonoglou, Hubert, Simonyan,
  Sifre, Schmitt, Guez, Lockhart, Hassabis, Graepel,
  et~al.]{schrittwieser2019mastering}
Schrittwieser, J., Antonoglou, I., Hubert, T., Simonyan, K., Sifre, L.,
  Schmitt, S., Guez, A., Lockhart, E., Hassabis, D., Graepel, T., et~al.
\newblock Mastering atari, go, chess and shogi by planning with a learned
  model.
\newblock \emph{Nature}, 588\penalty0 (7839):\penalty0 604--609, 2020.

\bibitem[Sermanet et~al.(2018)Sermanet, Lynch, Chebotar, Hsu, Jang, Schaal,
  Levine, and Brain]{sermanet2018time}
Sermanet, P., Lynch, C., Chebotar, Y., Hsu, J., Jang, E., Schaal, S., Levine,
  S., and Brain, G.
\newblock Time-contrastive networks: Self-supervised learning from video.
\newblock In \emph{2018 IEEE International Conference on Robotics and
  Automation (ICRA)}, pp.\  1134--1141. IEEE, 2018.

\bibitem[Silver et~al.(2017)Silver, van Hasselt, Hessel, Schaul, Guez, Harley,
  Dulac{-}Arnold, Reichert, Rabinowitz, Barreto, and
  Degris]{silver2016predictron}
Silver, D., van Hasselt, H., Hessel, M., Schaul, T., Guez, A., Harley, T.,
  Dulac{-}Arnold, G., Reichert, D.~P., Rabinowitz, N.~C., Barreto, A., and
  Degris, T.
\newblock The predictron: End-to-end learning and planning.
\newblock In Precup, D. and Teh, Y.~W. (eds.), \emph{Proceedings of the 34th
  International Conference on Machine Learning, {ICML} 2017, Sydney, NSW,
  Australia, 6-11 August 2017}, volume~70 of \emph{Proceedings of Machine
  Learning Research}, pp.\  3191--3199. {PMLR}, 2017.

\bibitem[Srinivas et~al.(2018)Srinivas, Jabri, Abbeel, Levine, and
  Finn]{srinivas2018universal}
Srinivas, A., Jabri, A., Abbeel, P., Levine, S., and Finn, C.
\newblock Universal planning networks: Learning generalizable representations
  for visuomotor control.
\newblock In \emph{International Conference on Machine Learning}, pp.\
  4732--4741. PMLR, 2018.

\bibitem[Stone et~al.(2021)Stone, Ramirez, Konolige, and
  Jonschkowski]{stone2021distracting}
Stone, A., Ramirez, O., Konolige, K., and Jonschkowski, R.
\newblock The distracting control suite--a challenging benchmark for
  reinforcement learning from pixels.
\newblock \emph{arXiv preprint arXiv:2101.02722}, 2021.

\bibitem[Stooke et~al.(2020)Stooke, Lee, Abbeel, and
  Laskin]{stooke2020decoupling}
Stooke, A., Lee, K., Abbeel, P., and Laskin, M.
\newblock Decoupling representation learning from reinforcement learning.
\newblock \emph{arXiv preprint arXiv:2009.08319}, 2020.

\bibitem[Tamar et~al.(2016)Tamar, Levine, Abbeel, Wu, and
  Thomas]{tamar2016value}
Tamar, A., Levine, S., Abbeel, P., Wu, Y., and Thomas, G.
\newblock Value iteration networks.
\newblock In Lee, D.~D., Sugiyama, M., von Luxburg, U., Guyon, I., and Garnett,
  R. (eds.), \emph{Advances in Neural Information Processing Systems 29: Annual
  Conference on Neural Information Processing Systems 2016, December 5-10,
  2016, Barcelona, Spain}, pp.\  2146--2154, 2016.

\bibitem[Tassa et~al.(2018)Tassa, Doron, Muldal, Erez, Li, de~Las~Casas,
  Budden, Abdolmaleki, Merel, Lefrancq, Lillicrap, and
  Riedmiller]{tassa2018dmc}
Tassa, Y., Doron, Y., Muldal, A., Erez, T., Li, Y., de~Las~Casas, D., Budden,
  D., Abdolmaleki, A., Merel, J., Lefrancq, A., Lillicrap, T., and Riedmiller,
  M.
\newblock Deepmind control suite.
\newblock January 2018.

\bibitem[Wahlstr{\"o}m et~al.(2015)Wahlstr{\"o}m, Sch{\"o}n, and
  Deisenroth]{wahlstrom2015pixels}
Wahlstr{\"o}m, N., Sch{\"o}n, T.~B., and Deisenroth, M.~P.
\newblock From pixels to torques: Policy learning with deep dynamical models.
\newblock \emph{arXiv preprint arXiv:1502.02251}, 2015.

\bibitem[Wang et~al.(2019)Wang, Ge, Lipton, and Xing]{wang2019learning}
Wang, H., Ge, S., Lipton, Z., and Xing, E.~P.
\newblock Learning robust global representations by penalizing local predictive
  power.
\newblock In \emph{Advances in Neural Information Processing Systems}, pp.\
  10506--10518, 2019.

\bibitem[Watter et~al.(2015)Watter, Springenberg, Boedecker, and
  Riedmiller]{watter2015embed}
Watter, M., Springenberg, J., Boedecker, J., and Riedmiller, M.
\newblock Embed to control: A locally linear latent dynamics model for control
  from raw images.
\newblock \emph{Advances in neural information processing systems},
  28:\penalty0 2746--2754, 2015.

\bibitem[Yan et~al.(2020)Yan, Vangipuram, Abbeel, and Pinto]{yan2020learning}
Yan, W., Vangipuram, A., Abbeel, P., and Pinto, L.
\newblock Learning predictive representations for deformable objects using
  contrastive estimation.
\newblock \emph{arXiv preprint arXiv:2003.05436}, 2020.

\bibitem[Yarats et~al.(2019)Yarats, Zhang, Kostrikov, Amos, Pineau, and
  Fergus]{yarats2019sample}
Yarats, D., Zhang, A., Kostrikov, I., Amos, B., Pineau, J., and Fergus, R.
\newblock Improving sample efficiency in model-free reinforcement learning from
  images.
\newblock October 2019.

\bibitem[Yarats et~al.(2021)Yarats, Kostrikov, and Fergus]{kostrikov2020image}
Yarats, D., Kostrikov, I., and Fergus, R.
\newblock Image augmentation is all you need: Regularizing deep reinforcement
  learning from pixels.
\newblock In \emph{International Conference on Learning Representations}, 2021.

\bibitem[Zhang et~al.(2020)Zhang, Lyle, Sodhani, Filos, Kwiatkowska, Pineau,
  Gal, and Precup]{zhang2020invariant}
Zhang, A., Lyle, C., Sodhani, S., Filos, A., Kwiatkowska, M., Pineau, J., Gal,
  Y., and Precup, D.
\newblock Invariant causal prediction for block mdps.
\newblock In \emph{International Conference on Machine Learning}, pp.\
  11214--11224. PMLR, 2020.

\bibitem[Zhang et~al.(2021)Zhang, McAllister, Calandra, Gal, and
  Levine]{zhang2020learning}
Zhang, A., McAllister, R.~T., Calandra, R., Gal, Y., and Levine, S.
\newblock Learning invariant representations for reinforcement learning without
  reconstruction.
\newblock In \emph{International Conference on Learning Representations}, 2021.

\end{thebibliography}
\bibliographystyle{icml2021}
\clearpage
\appendix
\onecolumn
\section{Experimental Details}

\subsection{Additional Ablation}
\begin{figure*}[ht]
    \vspace{-.25em}
    \begin{subfigure}[b]{15em}
        \centering
        \includegraphics[height=9em,width=15em]{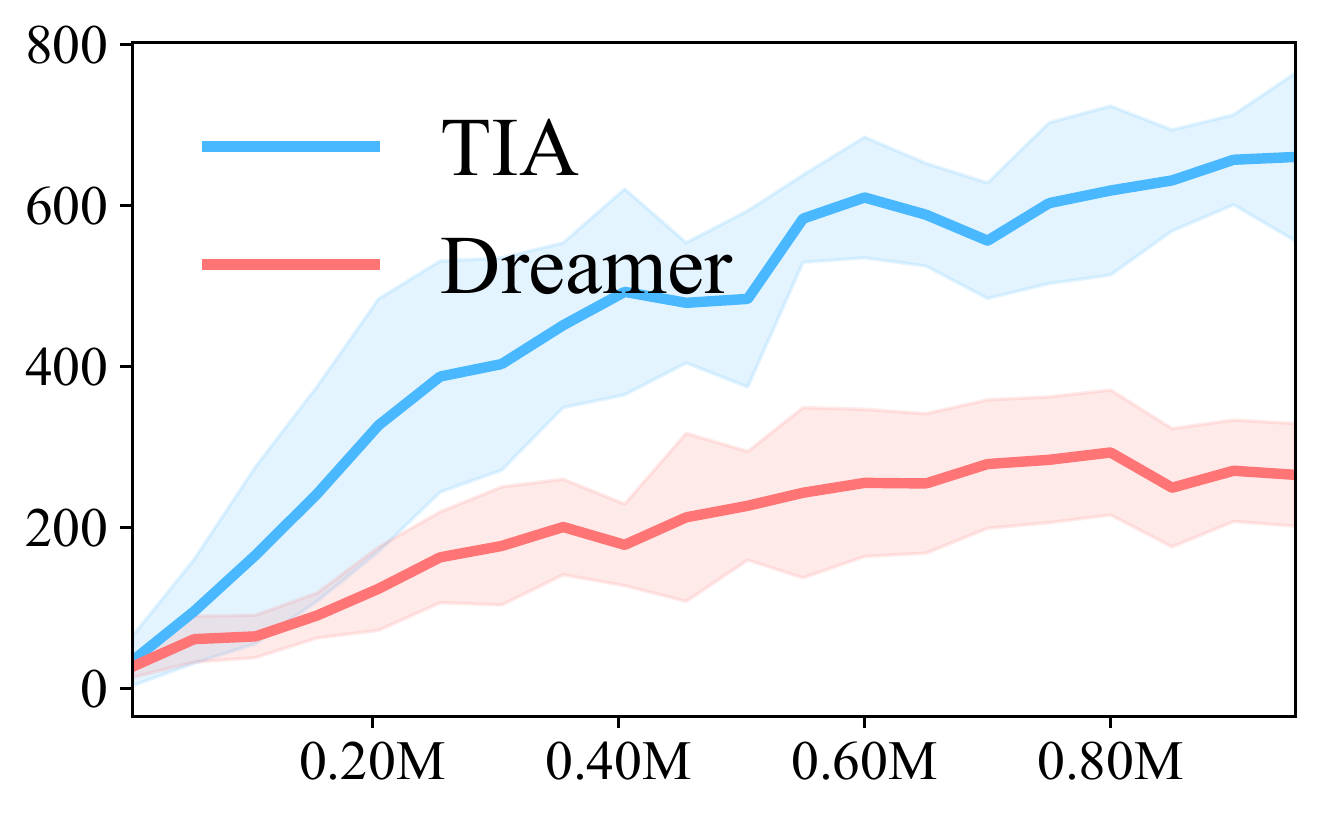}
        \caption{With 16 videos}
        \label{fig:16_videos}
    \end{subfigure}\hfill
    \begin{subfigure}[b]{17em}
        \centering
        \includegraphics[height=9em,width=17em]{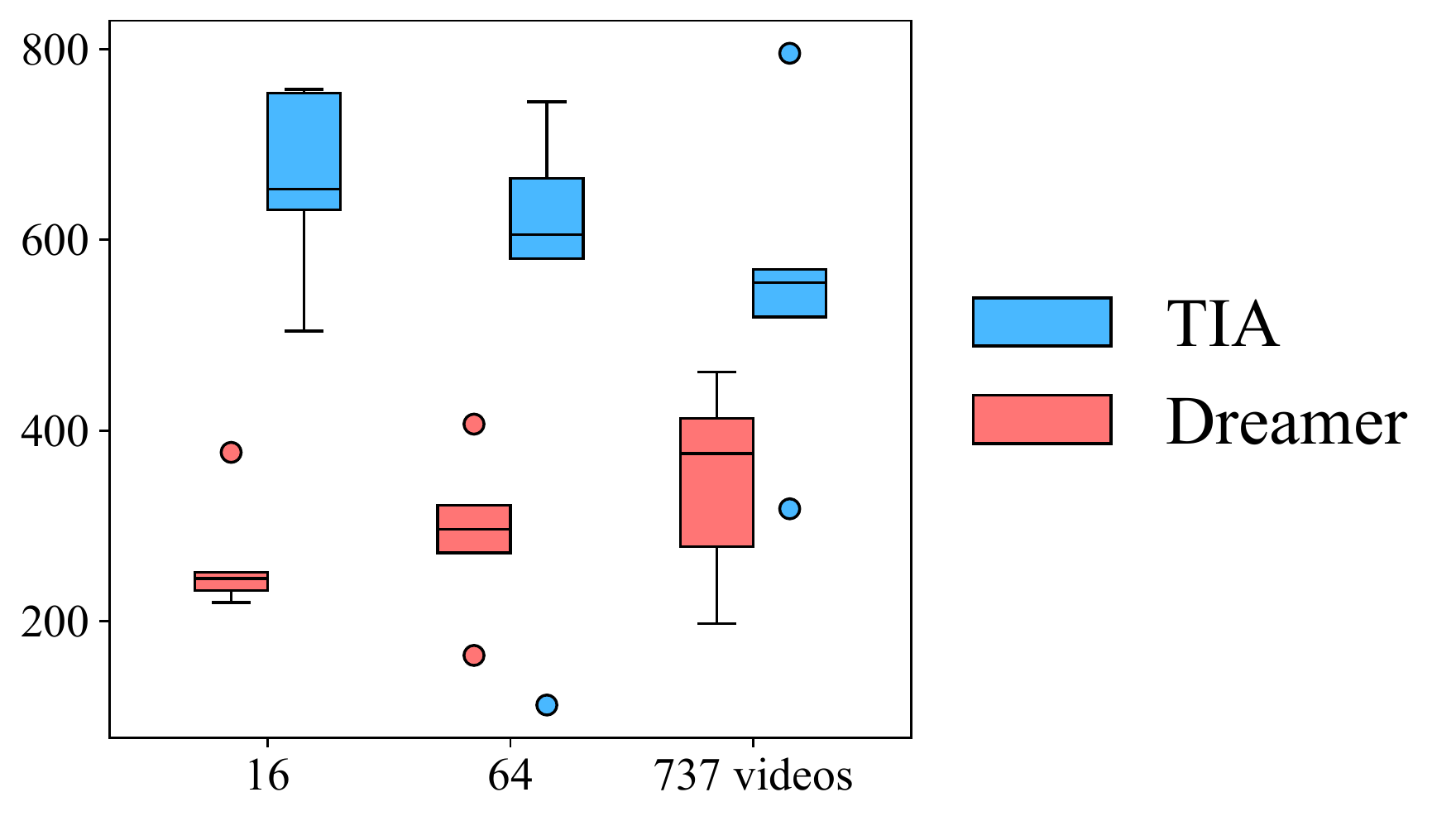}
        \caption{Variance from number of videos}
        \label{fig:box_plot}
    \end{subfigure}\hfill
    \begin{subfigure}[b]{15em}
        \centering
        \includegraphics[height=9em,width=15em]{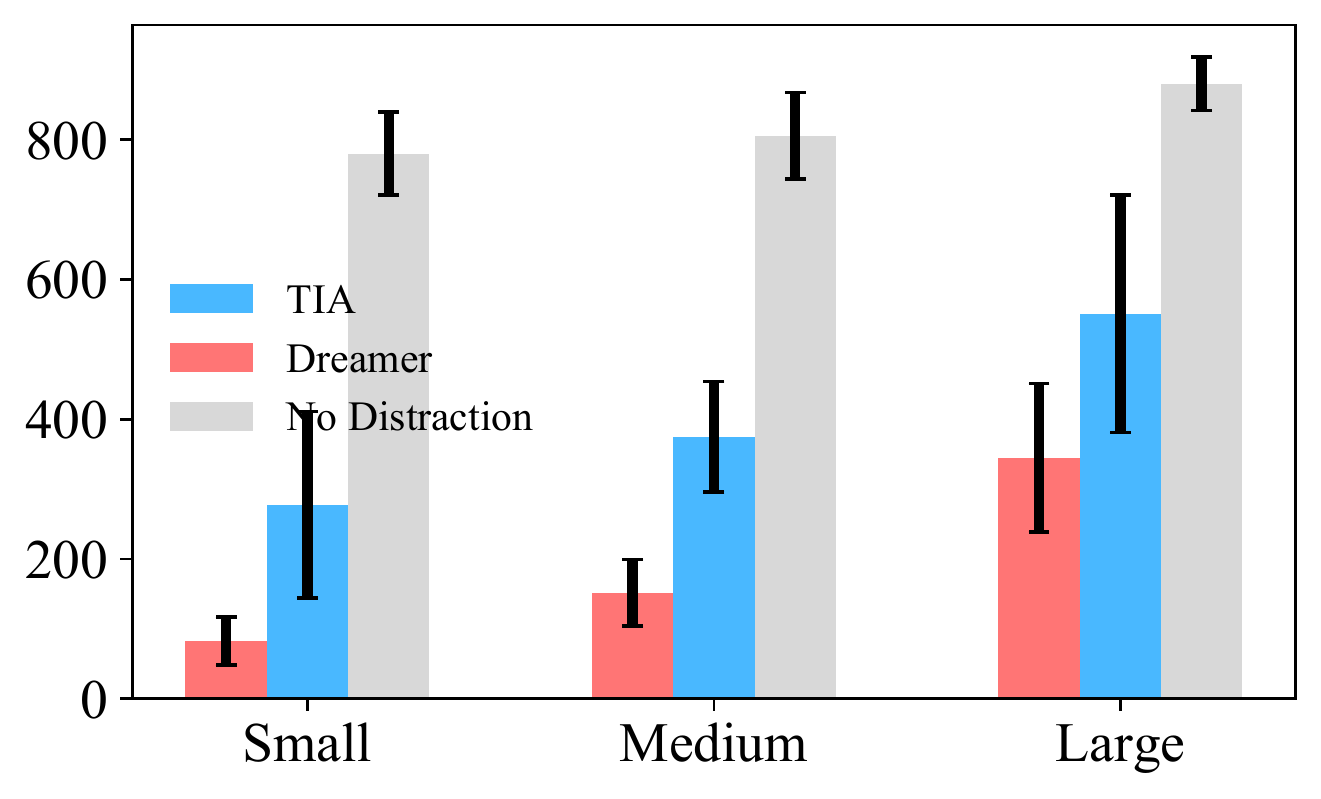}
        \caption{Varying model size}
        \label{fig:motivation_follow_up}
    \end{subfigure}
    \vspace{-1em}
    \caption{(a) With 16 videos used in the environment, the learning curve has much smaller variance, showing TIA significantly outperforms Dreamer. (b) Performance at 1M steps over the five seeds, using different number of videos in the environment. (c) TIA outperforms Dreamer irrespective of the model size at 1M steps.}
\end{figure*}

The learning curves for the DMC domains in \cref{fig:learning_manydmc} has large variance. We run a two-tailed test for performance at 1M steps, the results are: \([t\!=\!3.17,\,p\!=\!0.014]\), \([t\!=\!2.28,\,p\!=\!0.052]\), and \([t\!=\!2.21,\,p\!=\!0.061]\) on \textit{Hopper}, \textit{Walker} and \textit{Cheetah}, respectively. A major source of variance in the result is that the set of 737 videos in the natural video dataset are only sampled 1000 times in total (at the beginning of each episode). Therefore the model is constantly surprised by visually distinct videos (see outliers in \cref{fig:box_plot}), leading to large variance in the episodic reward for all algorithms considered. To control for this, we present additional results on \textit{Cheetah Run} that only uses 16 videos in \cref{fig:16_videos} and a comparison across \(16, 64, 737\) videos in a box plot (\cref{fig:box_plot}). The variance decreases significantly when this is controlled for, showing that TIA outperforms Dreamer. One interesting observation is that Dreamer performs worse with a smaller number of videos (potentially because it prioritizes memorization of the video background).

We further reproduce \cref{fig:model_capacity} while including TIA's performance in \cref{fig:motivation_follow_up}. TIA significantly outperforms Dreamer irrespective of the model size.

\subsection{Performance Gap When Learning In The Presence of Visual Distractions}\label{sec:perf_gap}

\begin{figure*}[h]
    \centering
    \includegraphics[width=\textwidth]{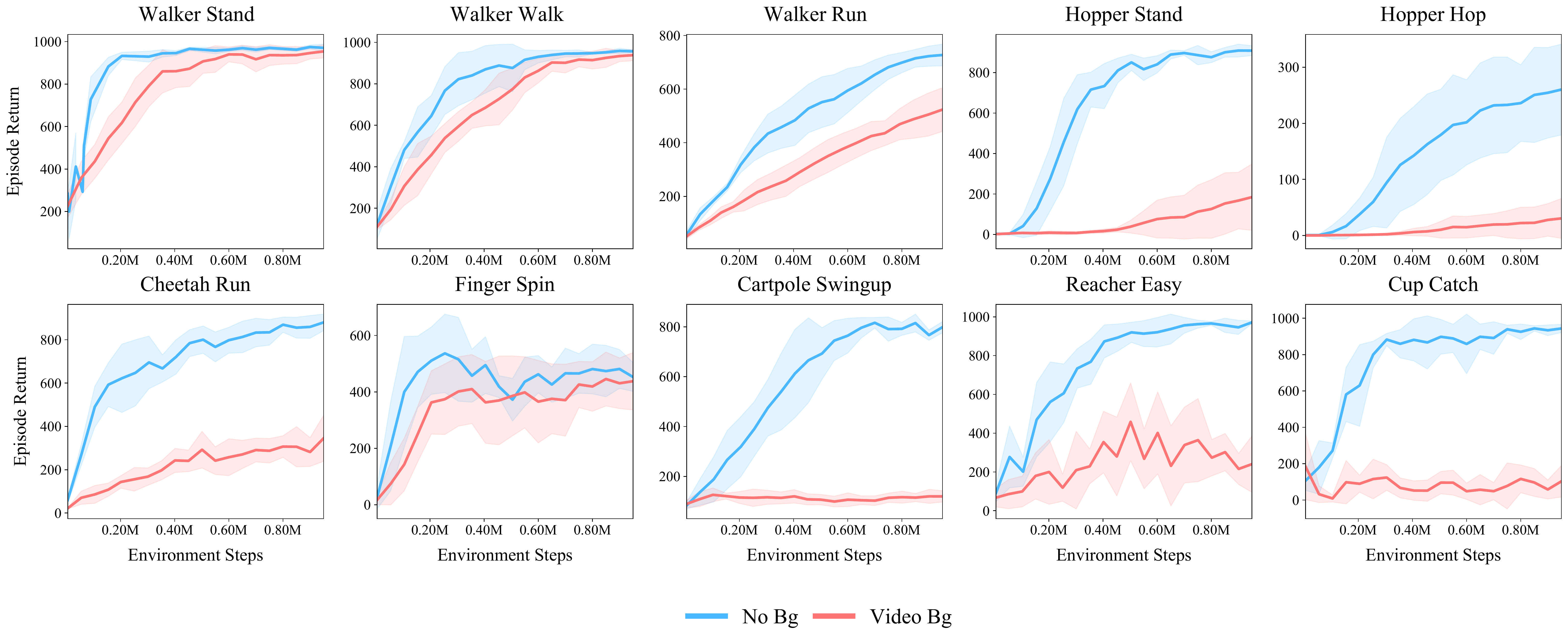}
    \vspace{-1.5em}
    \caption{\textbf{The data efficiency and performance gap} when learning with video distraction backgrounds. }%
    \label{fig:dmc_drop}
\end{figure*}

When complex visual distractors are present in the observations, state-of-the-art model-based agents struggle to maintain their original sample efficiency and asymptotic performance. We produce this gap in details with~\cref{fig:dmc_drop}, where we include seven additional (ten in total) domains from the DeepMind Control suite. In \textit{Finger Spin} the gap almost disappears due to poor performance of the model-based agent in the clean environment -- making the domain ill-suited for testing our proposal. Note that the maximum episodic return on these tasks is calibrated to 1000~\cite{tassa2018dmc}. We additionally provide the ratio for the number of pixels that are replaced by the background video for selected domains as a proxy for how much visual distraction is introduced: Walker 63\%, Hopper 77\%, Cheetah: 83\%.

Our intent with this paper and the proposal, learning \textit{Task Informed Abstractions}, is to close this gap such that model-based agents can retain their performance even when learning in the presence of complex visual distractions.

\subsection{Bridging the Gap by Learning Task Informed Abstractions}
We produce the complete result on DeepMind Control Suite, including the seven additional tasks from above. We set up the experiments by matching the total number of parameters so that each one of TIA's two models is only half in size as the single world model from Dreamer. This comparison is a disadvantage to our method because the smaller task model runs the risk of being too small to capture the set of task-relevant features in its entirety.
\begin{figure*}[h]
    \centering
    \includegraphics[width=\textwidth]{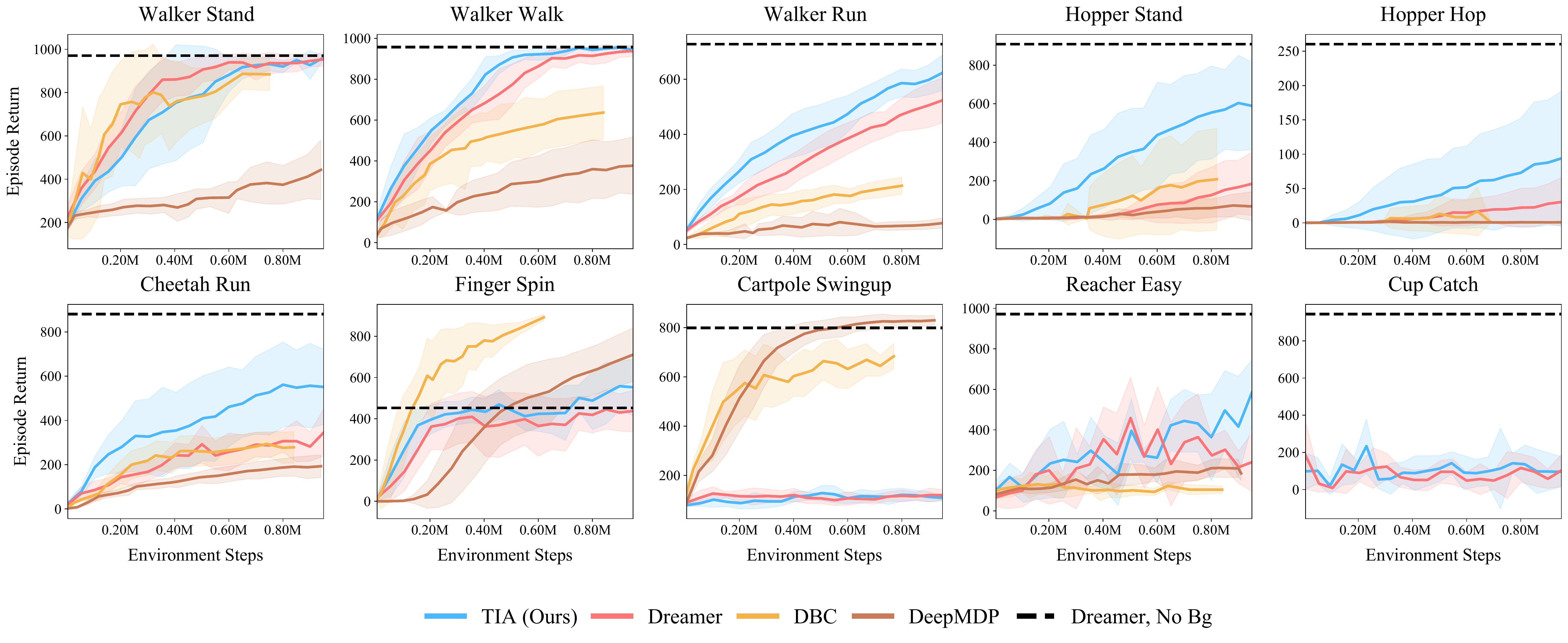}
    \caption{\textbf{Performance of Task Informed Abstractions} on ten DeepMind control domains. \citealt{zhang2020learning} did not evaluate DBC/DeepMDP on \textit{Cup Catch} so we only report results of TIA and Dreamer for \textit{Cup Catch}.}
    \label{fig:dmc_all}
\end{figure*}%

Despite of this disadvantage in model capacity, TIA is able to reduce the gap on \textit{Walker Run}, \textit{Hopper Hop}, and \textit{Reacher Easy} while making the gap significantly smaller on \textit{Hopper Stand} and \textit{Cheetah Run} (see~\cref{fig:dmc_drop}). On \textit{Walker Stand} and \textit{Walker Walk}, the gap is small to begin with. Therefore our method cannot bring much benefit. Domains such as \textit{Cartpole Swingup} and \textit{Cup Catch} are devastatingly hard for Dreamer to master under visual distraction due to the partial-observable/small object manipulation nature of the domains. To conclude, since our implementation of TIA is a Dreamer with an additional capability of ignoring task-irrelevant features, it would not perform worse than Dreamer. However, tasks that are inherently hard for model-based methods would remain hard for TIA. 

\subsection{Understanding Failure Cases}

\begin{figure*}[h]
\providecommand{\width}{}
\renewcommand{\width}{.25\textwidth}
\begin{subfigure}[t]{0.16\textwidth}
\includegraphics[height=215pt]{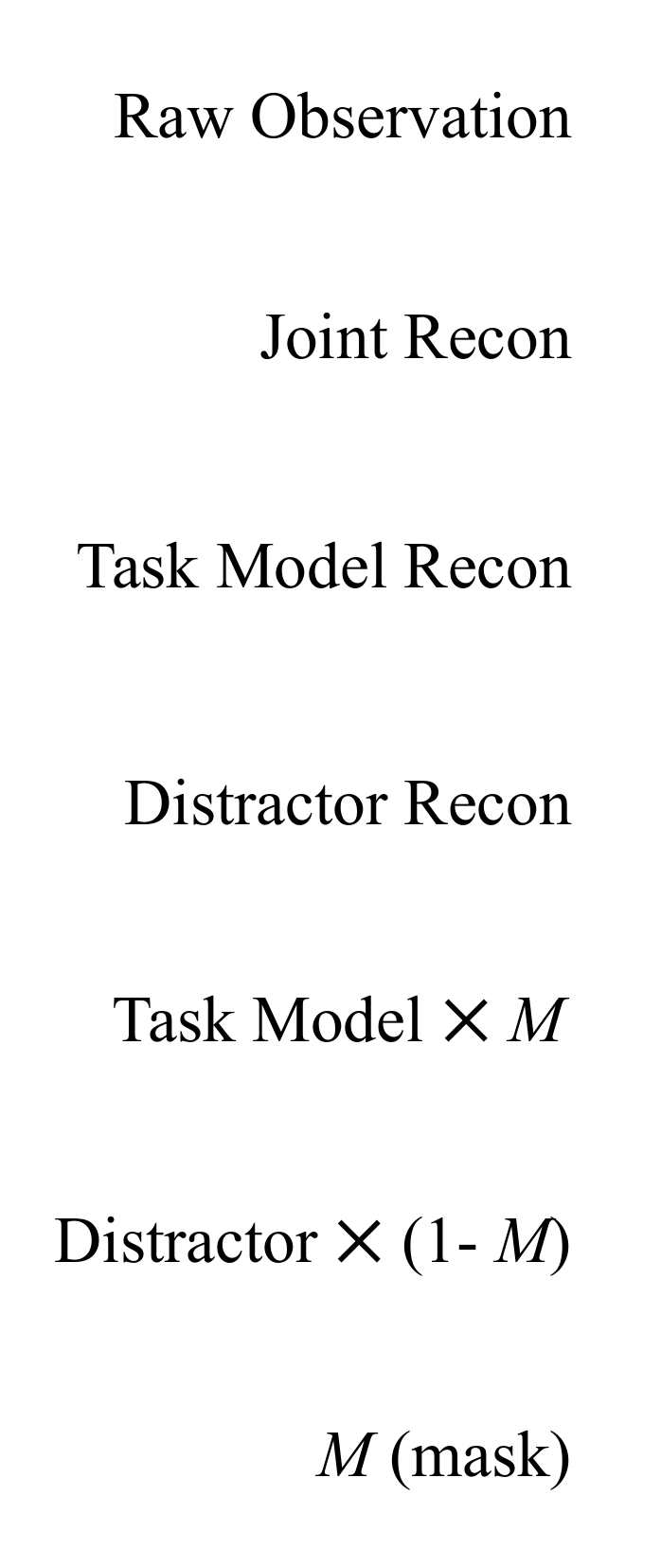}
\label{fig:rendering_many_dreamer}
\end{subfigure}\hfill%
\begin{subfigure}[t]{\width}
\includegraphics[height=215pt]{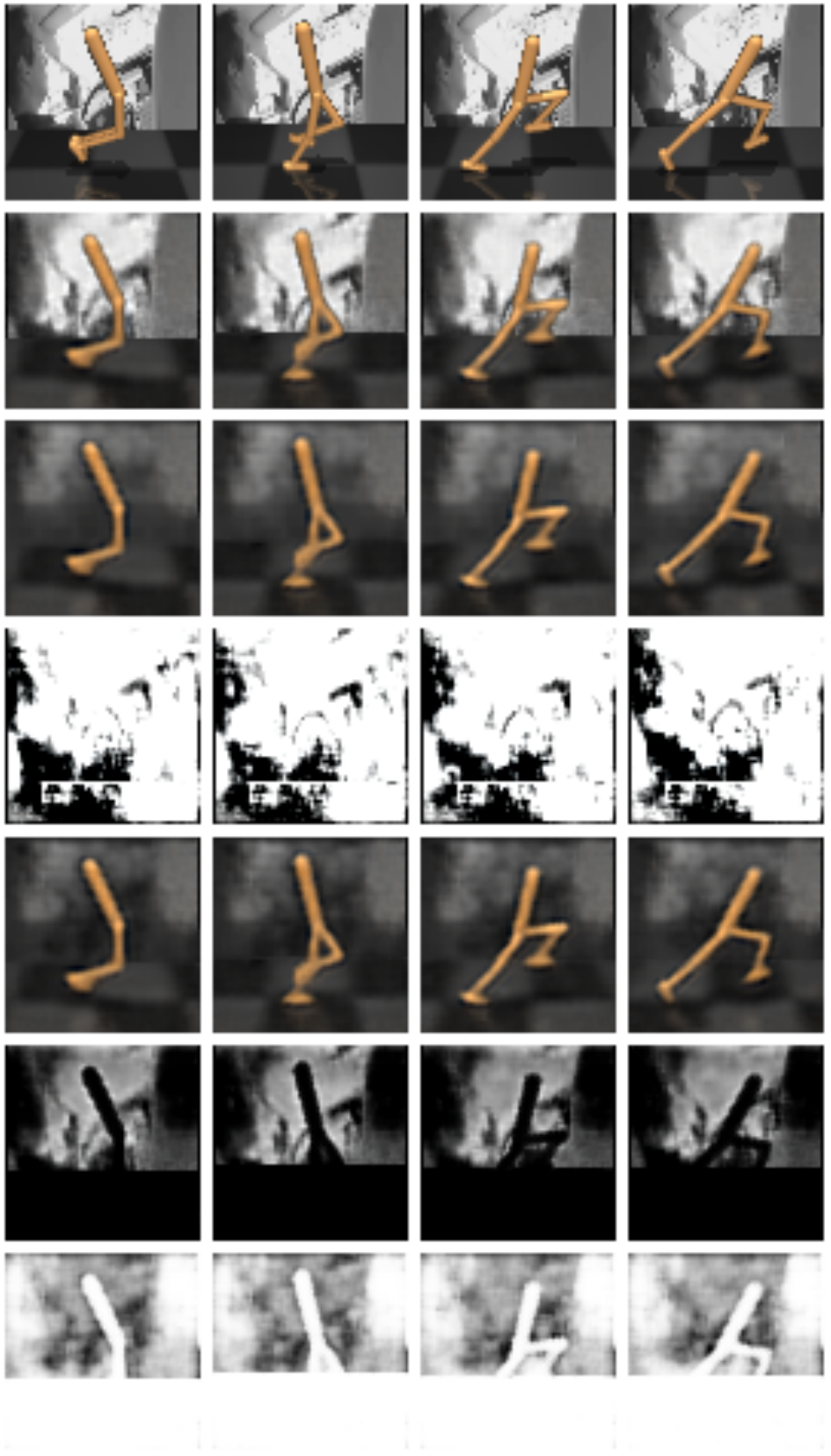}
\caption{Success}\label{fig:failure_success}
\end{subfigure}\hfill%
\begin{subfigure}[t]{\width}
\includegraphics[height=215pt]{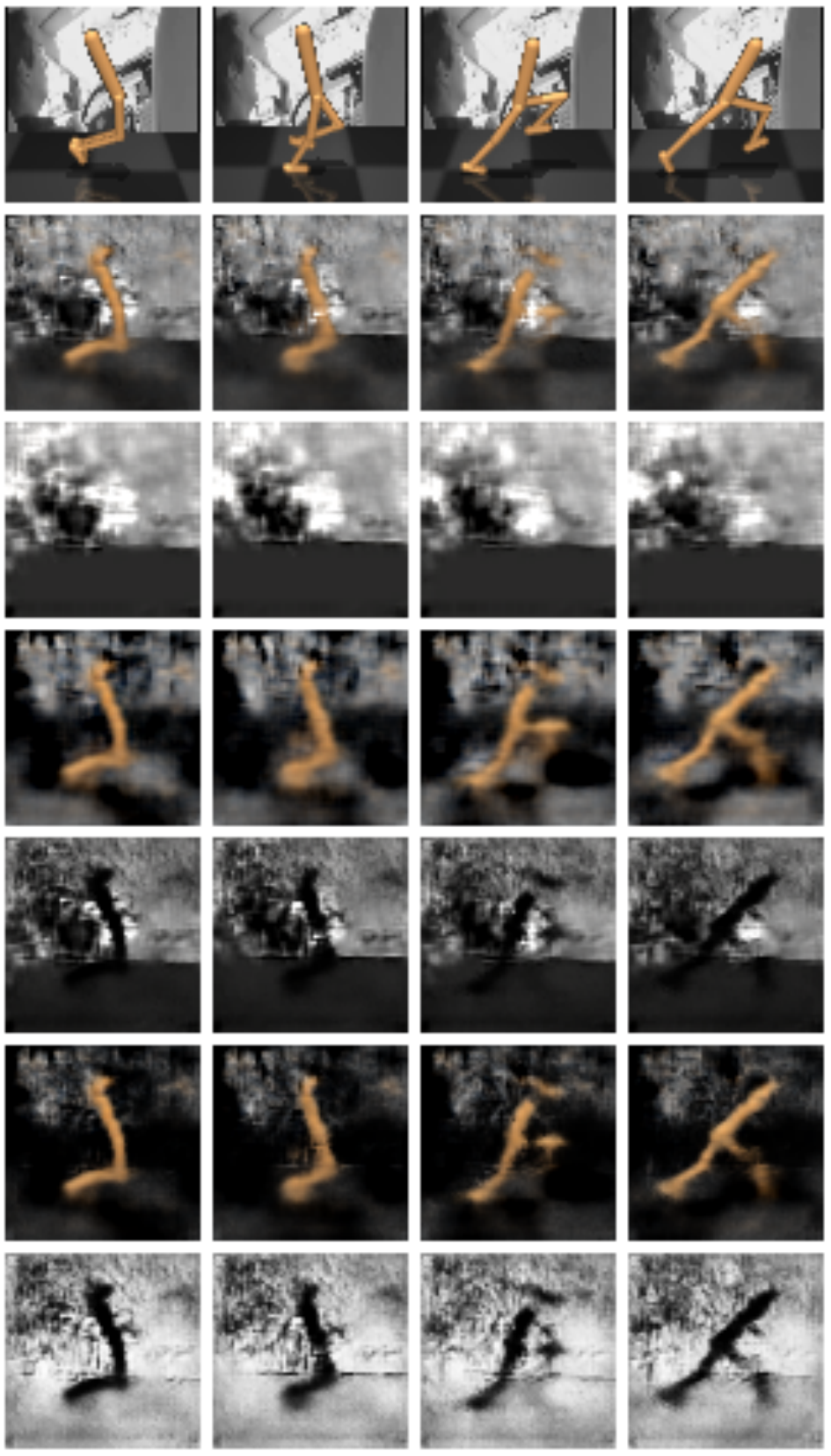}
\caption{Failure Type 1: The distractor model takes over in reconstruction}\label{fig:failure_1}
\end{subfigure}\hfill%
\begin{subfigure}[t]{\width}
\includegraphics[height=215pt]{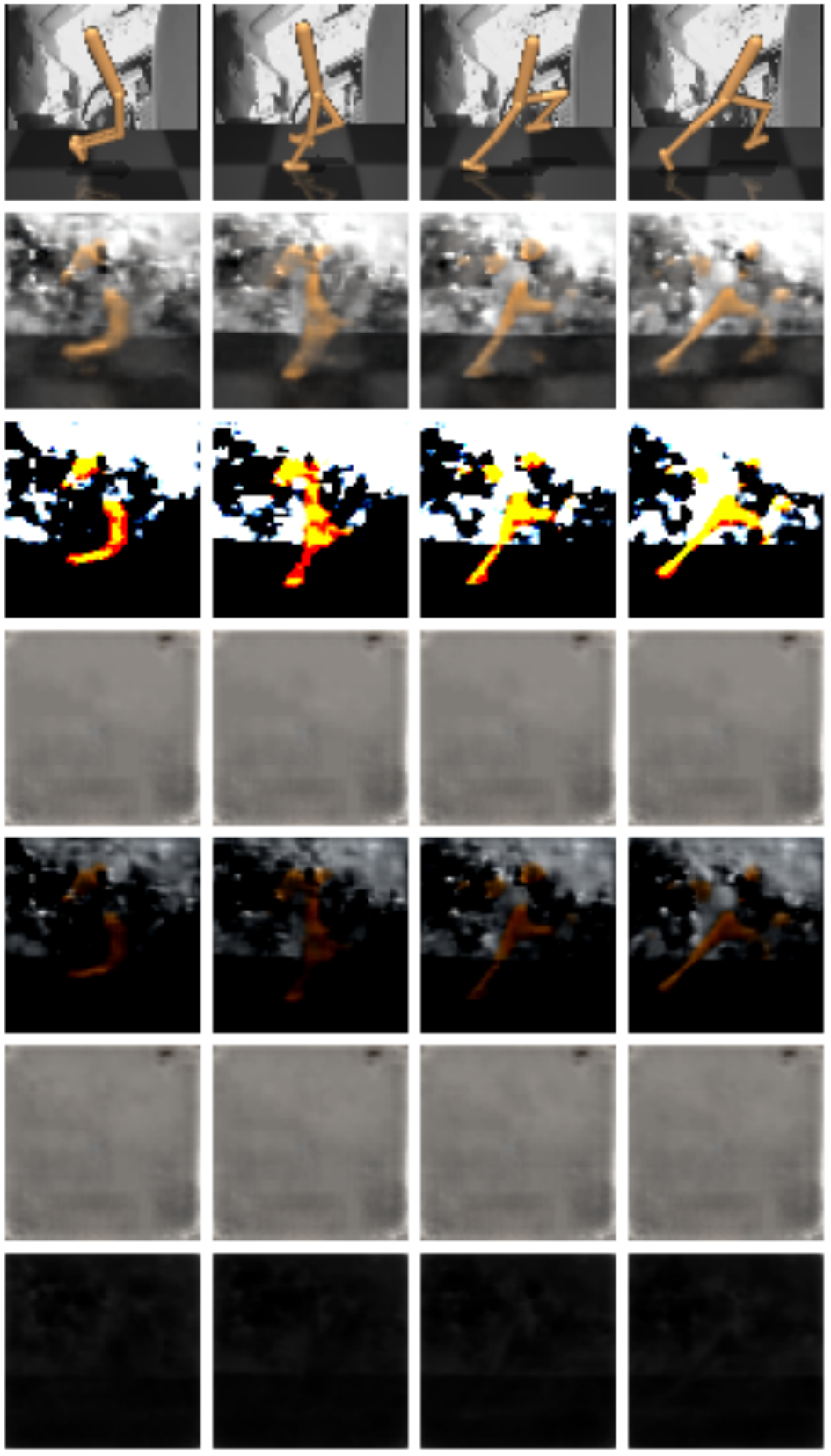}
\caption{Failure Type 2: The task model takes over in reconstruction}\label{fig:failure_2}
\end{subfigure}%
\caption{Detailed rendering from one successfully trained walker agent, and two failure cases of type 1 and type 2. In type 1 failure case, the distractor model takes over the entire reconstruction, rendering the task model ineffective for policy learning (random policy). In the type 2 failure case, the task model attempts to capture all factors of variation by itself, thus failing to perfectly reconstruct the image, also leading to sub-par policy performance in the lower 400.%
}
\label{fig:failure_rendering}
\vspace*{-1ex}
\end{figure*}

In Section $4.5$, we provide the principled approach to tuning the hyperparameters $\lambda_{\mathrm{Radv}}$ and $\lambda_{\mathrm{Os}}$, where we balance these two terms to avoid either one of the two models taking over the entire reconstruction. We label these two extreme cases, where either the distractor model or the task model takes over, as type I and type II. We provide detailed renderings of these failure cases below~\cref{fig:failure_rendering} in comparison to a successfully learned world model. 

The first column in ~\cref{fig:failure_success} shows a successfully trained agent whose task model is able to perfectly reconstruct the walker agent. In type I failure mode (\cref{fig:failure_1}), the distractor model would take over the entire reconstruction, causing the task model to lose its grasp on the task-relevant features. This would prevent the policy from learning any useful behavior, which makes the collected reward practically random. With random rewards, the term \(\mathcal J_\mathrm{Radv}\) is ineffective at dissociating task-relevant features, causing training to fail. The policy performance under this scenario is usually close to that of a random policy. When this happens, we want to increase  $\lambda_{\mathrm{Radv}}$ to dissociate the distractor model from the reward, or decrease \(\lambda_\mathrm{Os}\) that weighs the distractor reconstruction term.

In the type II failure case (see~\cref{fig:failure_2}), the task model takes over the reconstruction. In this case, the model degenerates into Dreamer without separating out the task-irrelevant features, and the performance is close to a dreamer agent with a smaller model. In this case, we would increase the weight \(\lambda_\mathrm{Os}\) on the distractor reconstruction to encourage it to learn more features. 

By tuning these two parameters $\lambda_{\mathrm{Radv}}$ and $\lambda_{\mathrm{Os}}$, we were able to avoid these two failure modes and stabilize training on a wide variety of domains. A future direction would be to tune these to parameters automatically using the signals mentioned in Section $4.5$.

\section{Model Architecture Details}

\paragraph{Model Detals and Sizes}
For fair comparisons, on DMC and ManyWorld we match the total number of parameters of our two world models combined with that of a single, large Dreamer model. The total number of parameters is 10 million on DMC and 1 million on ManyWorld. We divide the two models equally in size, with an additional image reconstruction head for the distractor model.  \textit{Dreamer-Inverse} has the same size for all model components as the Dreamer model, except it replaces the deconvolution head with an action prediction head for learning the inverse dynamics.  We scale the model size by changing the width of each layer in the networks without changing the overall architecture or the depth of the networks. All other hyperparameters such as learning rates are kept the same as \cite{hafner2019dream}.

On the Arcade Learning Environments, we compare against the state-of-the-art on this domain, DreamerV2 \cite{hafner2020mastering}, which uses a world model of 20 million trainable parameters. We made the task model the same size as the original implementation while adding a smaller distractor model which contains 12 million parameters. A key difference between Dreamer and DreamerV2 is that the latter has an additional prediction head for the discount factor \(\gamma_t\) besides the standard reward prediction head. This discount factor head plays an instrumental role in allowing DreamerV2 to solve Atari games. Therefore we additionally dissociate the distractor model from information about the discount factor by adding an additional adversarial prediction loss. We use the same scale for the discount factor $\gamma$ as that for reward: $\lambda_{\gamma \mathrm{adv}} = \lambda_{\mathrm{Radv}}$.  All other hyperparameters such as learning rates are kept the same as \cite{hafner2020mastering}.

We use an input size of $32\!\times\! 32\!\times\! 3$ in ManyWorld, $64\!\times\!64\!\times\!3$ in DeepMind Control Suite, and $64\!\times\!64\!\times\!1$ in Atari games. We use grayscale for the natural video backgrounds, the same as previous work \cite{zhang2020learning}.

\paragraph{Hyperparameters} For fair comparisons, we did not tweak existing hyperparameters from Dreamer and used identical settings as \citet{hafner2019dream} and \citet{hafner2020mastering}. Our reward-dissociation scheme introduces two new hyperparameters \(\lambda_\mathrm{Radv}\) and \(\lambda_\mathrm{Os}\). We scale the reward dissociation loss via \(\lambda_\mathrm{Radv}\) such that the term matches reconstruction losses in magnitude. For this reason, the differences in scale in \cref{tab:lambdas} mostly reflect the differences in input image sizes. We tweaked \(\lambda_\mathrm{Os}\) to stabilize training. Detailed settings for each domain are in \cref{tab:lambdas}.

\begin{table}[h]
\centering
\caption{Hyperparameters}\label{tab:lambdas}
\begin{tabular}{lll}
\toprule
  Domain and Task       & $\lambda_{\mathrm{Radv}}$ & $\lambda_{\mathrm{Os}}$ \\ 
\midrule
ManyWorld, 1 Distractor & 600.0                     & 2.0            \\ 
ManyWorld, 2 Distractor & 150.0                     & 2.0            \\ 
Cartpole Swingup        & 30k                   & 2.0 
\\
Hopper Stand            & 30k                   & 2.0                  \\ 
Hopper Hop              & 30k                   & 2.0                  \\ 
Cheetah Run             & 20k                   & 1.5                   \\ 
Walker Run              & 20k                   & 0.25                  \\ 
Walker Walk             & 20k                   & 0.25                 \\
Walker Stand            & 20k                   & 0.25                 \\
Finger Spin             & 30k                   & 2.5                     \\ 
Reacher Easy            & 20k                   & 1.0
\\
All ATARI games         & 2k                    & 1.0                \\
\bottomrule
\end{tabular}
\end{table}

\section{Transfer to Novel Distractions}

The task informed abstraction we introduce in this paper improves learning when distractions are present. To adapt to out-of-distribution scenarios unseen during training, additional architectural changes that reject distracting image features on the fly may be required. To provide a baseline and intuitions for this future direction, we evaluate how well existing agents perform under this type of domain shift. In \cref{tab:transfer_dmc} we take agents that are trained (1) without video background, (2) with background videos from the \textit{driving car} class or (3) with white noise backgrounds,  and evaluate against background videos from a different class, \textit{walking the dog} (labeled as \textit{transfer}, see~\cref{tab:transfer_dmc}). 
\begin{table*}[h]
\caption{\textbf{DeepMind Control Transfer Performance}\quad transfer to the video class \textit{walking the dog} as background}\label{tab:transfer_dmc}
\centering
\label{table:DMC_transfer}
\begin{tabular}{l|l|lllll}
\toprule
\multicolumn{2}{r|}{Training Condition}   & Drmr, No Bg    & Drmr, Video         & TIA, Video         & Drmr, Noise       & TIA, Noise        \\
\midrule
\multirow{2}{*}{Hopper Stand} & In-domain & $906.8 \pm 29.3$  & $183.8 \pm 162.1$   & $596.4 \pm 234.1$  & $769.7 \pm 205.4$ & $744.8 \pm 75.8$  \\
                              & Transfer  & \,$18.1 \pm 12.1$ & $186.2 \pm 142.2$   & $629.4 \pm 231.5$  & $357.4 \pm 206.7$ & $354.9 \pm 136.9$ \\
\midrule
\multirow{2}{*}{Walker Run}   & In-domain & $728.2 \pm 37.8$  & $520.2 \pm 84.4$    & $625.3 \pm 64.7$   & $737.6 \pm 26.7$  & $696.9 \pm 43.9$  \\
                              & Transfer  & $127.4 \pm 34.0$  & $530.9 \pm 76.9$    & $645.3 \pm 78.4$   & $341.1 \pm 108.9$ & $345.5 \pm 138.0$ \\
\midrule
\multirow{2}{*}{Cheetah Run}  & In-domain & $876.3 \pm 36.0$  & $325.7 \pm 96.6$    & $556.6 \pm 167.7$  & $754.9 \pm 67.0$  & $734.2 \pm 163.4$ \\
                              & Transfer  & \,$21.1 \pm 7.7$  & $312.6 \pm 115.7$   & $557.4 \pm 194.6$  & $227.0 \pm 75.1$  & $309.5 \pm 233.2$ \\
\bottomrule
\end{tabular}
\end{table*}

The Dreamer agent trained with no background distraction fails to transfer its performance when background videos are introduced at test time, which is expected. In the second experiment, we train both Dreamer and TIA with video background but test using videos from a different category. Dreamer did not learn as well as TIA as indicated by its poor performance in the training environments, but both methods retain their training performance post-transfer, unaffected by the change in the background video. As a control, we also train both methods using white noise as the background. The training and transfer performance are both identical between the two methods, and the transfer performance is worse than performance on white noise.

We additionally evaluate ManyWorld agents that are trained with (1) no distractor, (2) one distractor, or (3) two distractors, with an additional distractor (three distractors, see~\cref{tab:transfer_many}). 
\begin{table*}[h]
\caption{\textbf{ManyWorld Transfer Performance}\quad transfer to three distraction blocks.}\label{tab:transfer_many}
\centering
\label{table:DMC_transfer}
\begin{tabular}{l|l|lllll}
\toprule
\multicolumn{2}{r|}{Training Condition} & Drmr, 0             & Drmr, 1            & Drmr, 2          & TIA, 1         & TIA, 2 \\
\midrule
\multirow{2}{*}{ManyWorld}  & In-domain & $246.0 \pm 3.5$     & $242.9 \pm 5.4$    & $217.4 \pm 29.3$ & $246.1 \pm 1.7$  & $245.8 \pm 1.8$ \\
                            & Transfer  & $192.6 \pm 18.9$    & $198.4 \pm 27.4$   & $192.0 \pm 35.1$ & $185.7 \pm 21.4$ & $233.4 \pm 6.5$ \\
\bottomrule
\end{tabular}
\end{table*}

Both results show that while TIA learns better from cluttered scenes, a mechanism to reject unseen backgrounds at decision time is required to transfer successfully. This points to the incorporation of attention as a potential avenue for future work.
\end{document}